\documentclass{article}

\usepackage[preprint]{neurips_2026}
\usepackage[utf8]{inputenc} 
\usepackage[T1]{fontenc}    
\usepackage{hyperref}       
\usepackage{url}            
\usepackage{booktabs}       
\usepackage{amsfonts}       
\usepackage{nicefrac}       
\usepackage{microtype}      
\usepackage{xcolor}         
\usepackage{graphicx}
\usepackage{amsmath}
\usepackage{amsfonts}
\usepackage{subcaption}
\usepackage{caption}
\usepackage{float}
\usepackage{mathtools}
\usepackage{extpfeil} 
\usepackage{natbib}
\usepackage{wrapfig}
\usepackage{overpic}
\usepackage{xspace}
\usepackage{enumitem}
\usepackage{xurl}

\usepackage{multirow}
\usepackage{multicol}
\usepackage{siunitx}
\usepackage{bm}
\usepackage{listings}
\usepackage{tikz}
\usepackage{tabularx}   
\usepackage{tabularray} 

\usepackage{authblk}
\usepackage{bbm}

\title{TGPO: Trace-Guided Policy Optimization for Robot Task Planning via Verifiable Subgoal Generation}

\author{
Zhihong Liu$^{1,3,*}$ \quad
Yang Li$^{1,4,*}$ \quad
Renming Huang$^{1,2}$ \quad
Chendong Zeng$^{1,2}$ \quad
Cewu Lu$^{1,2}$ \quad
Panpan Cai$^{1,2,\dagger}$\\
$^{1}$Shanghai Innovation Institute \quad
$^{2}$Shanghai Jiao Tong University \quad
$^{3}$Xi'an Jiao Tong University \quad
$^{4}$Nankai University\\
$^{*}$Equal contribution \quad
$^{\dagger}$Corresponding author
}
\hypersetup{
    colorlinks=true,
    citecolor=blue,
    linkcolor=blue,
    urlcolor=blue,
    bookmarks=true
}
\usepackage[noend]{algpseudocode}
\usepackage{adjustbox}
\usepackage{algorithm}
\usepackage{amsfonts}
\usepackage{amsmath}
\usepackage{amssymb}
\usepackage{array, booktabs, makecell}
\usepackage{babel,blindtext}
\usepackage{braket}
\usepackage{graphicx}
\usepackage{multicol}
\usepackage{multirow}
\usepackage{siunitx}
\usepackage{subcaption}
\usepackage{tabularx}
\usepackage{times}
\usepackage{varwidth}
\usepackage{xspace}
\usepackage{verbatim}
\algrenewcommand\algorithmicindent{1.0em}%

\makeatletter
\newcommand\fs@spaceruled{\def\@fs@cfont{\bfseries}\let\@fs@capt\floatc@ruled
  \def\@fs@pre{\vspace{0.6\baselineskip}\hrule height.8pt depth0pt \kern2pt}%
  \def\@fs@post{\kern2pt\hrule\relax}%
  \def\@fs@mid{\kern2pt\hrule\kern2pt}%
  \let\@fs@iftopcapt\iftrue}
\makeatother

\newcommand{\gptfive}{GPT-5\xspace}

\newcommand{\geminithreepro}{Gemini-3.0-pro\xspace}
\newcommand{\algsayplan}{SayPlan\xspace}
\newcommand{\algdelta}{DELTA\xspace}

\newcommand{\ourgrpo}{TGPO\xspace}

\newcolumntype{Y}{>{\centering\arraybackslash}X}

\usepackage{xcolor}

\newcommand{\secref}[1]{Section~\ref{#1}}

\renewcommand{\eqref}[1]{Eq. (\ref{#1})}
\newcommand{\figref}[1]{Fig.~\ref{#1}}

\newcommand{\eg}{\textit{e.g.}}

\begin{document}

\maketitle

\begin{abstract}
Robot task planning in real-world environments requires mapping abstract natural language instructions to executable action sequences under long horizons and complex constraints. While large language models (LLMs) provide strong commonsense reasoning, they often fail to generate reliable and feasible plans. In contrast, symbolic planners ensure feasibility and optimality but require well-specified goals and cannot directly interpret high-level human intent.
We formulate robot task planning as learning to generate \emph{verifiable subgoals} in the Planning Domain Definition Language (PDDL), bridging language understanding and symbolic planning. To address the challenges of sparse and noisy supervision, we propose Trace-Guided Policy Optimization (TGPO), a reinforcement learning framework that improves structured subgoal generation through (i) verifier-grounded rewards, (ii) external correction of intermediate reasoning traces, and (iii) constrained policy updates that incorporate corrected traces into training.
We evaluate TGPO on large-scale household planning tasks with long horizons, abstract instructions, and complex constraints. TGPO significantly outperforms prompting-based and reinforcement learning baselines, with the largest gains on abstract tasks. Furthermore, TGPO integrates naturally with symbolic planners and language-conditioned executors, enabling robust long-horizon planning and combinatorial generalization in realistic environments.\footnote{Project page: \url{https://tgpo2026.github.io/TGPO/}.}
\end{abstract}

\section{Introduction}\label{sec:intro}

Long-horizon planning from natural language remains a fundamental challenge for robotic systems. 
Real-world instructions are often abstract, underspecified, and require reasoning over extended horizons under complex constraints~\cite{li2023behavior, shridhar2020alfred}. 
While large language models (LLMs)~\cite{jaech2024openai, yang2025qwen3,Guo_2025} exhibit strong commonsense reasoning, they struggle to produce reliable and executable plans~\cite{yang2025embodiedbench, li2024embodied,huang2025limit}. 
This failure stems from several factors: (i) lack of explicit constraint modeling, leading to infeasible or inconsistent action sequences; 
(ii) compounding errors over long autoregressive generations, where early mistakes propagate and invalidate downstream steps; and 
(iii) weak grounding to environment state, especially in large-scale scenes. 
As a result, planning performance degrades sharply as task complexity and horizon increase.
In contrast, symbolic planners~\cite{aeronautiques1998pddl,erol1994umcp, garrett2021integrated} provide guarantees on feasibility and optimality, but require well-specified symbolic goals and cannot directly interpret high-level human intent. 
Bridging language understanding and symbolic reasoning is therefore central to scalable robot planning.

A common paradigm for integrating LLMs with symbolic planning~\cite{LLM+P_1, DELTA_11, Nl2plan_7,LLM-DP_4} is to have LLMs translate a language instruction into a symbolic goal specification, e.g., in Planning Domain Definition Language (PDDL)~\cite{PDDL_1}, which is then solved by a classical planner~\cite{fast_downward}. 
While effective for simple tasks, directly generating a single holistic goal becomes brittle as task complexity increases. 
Such goal specifications often fail to capture temporal dependencies among subtasks and omit key task elements implicit in the instruction; solving the planning problem can become prohibitively expensive or even impossible. 
A natural remedy is to decompose the task into intermediate subgoals~\cite{kwon2025fast,bai2024twostep}. 
Subgoals make key task elements and their temporal dependencies more explicit, and reduce solver complexity by breaking a long-horizon problem into smaller, modular subproblems. 
However, decomposition shifts the bottleneck to the \emph{formulation} of subgoals: the model must map an abstract instruction to a sequence of grounded, composable subgoals that are individually feasible, consistently chained without conflict or redundancy, and jointly sufficient to satisfy the implicit task intent.

In this work, we propose to \emph{learn} subgoal formulation as a structured policy. 
We introduce a foundation model that generates \emph{verifiable subgoals}, which, when coupled with a symbolic planner, enables scalable robot task planning in large-scale environments with long horizons, complex constraints, and abstract instructions. 
Our key idea is to directly optimize the policy for producing subgoal sequences that are feasible, consistent, and sufficient for task completion, rather than relying on prompting or post-hoc correction.
We formulate this as a reinforcement learning problem, where the policy is trained to maximize verifier-grounded rewards reflecting subgoal feasibility and overall task success. 
However, this learning problem is inherently challenging: reward signals are \emph{sparse}, since small errors in subgoal formulation often invalidate entire plans; \emph{noisy}, as satisfaction of implicit task intent is difficult to reliably evaluate; and \emph{delayed}, making credit assignment over intermediate subgoals particularly difficult. 

To address these challenges, we propose \textbf{Trace-Guided Policy Optimization (TGPO)}, a reinforcement learning algorithm for structured subgoal generation. 
TGPO is designed to tackle the key difficulties of sparse, noisy, and delayed feedback in subgoal formulation by introducing structured learning signals at the level of intermediate reasoning. 
Specifically, TGPO incorporates three components: 
(i) a \emph{verifier-grounded reward} that provides outcome-level supervision based on subgoal feasibility and task completion; 
(ii) \emph{trace-guided refinement}, where an auxiliary model identifies and corrects errors in intermediate reasoning traces, providing fine-grained guidance beyond scalar rewards; and 
(iii) \emph{constrained policy optimization}, which integrates corrected traces into training via token-level constraints while preserving autoregressive generation. 
Together, these mechanisms transform sparse and delayed rewards into dense, structured supervision, significantly improving exploration efficiency and credit assignment over long-horizon reasoning processes.

We evaluate TGPO on a diverse suite of long-horizon planning tasks spanning varying levels of abstraction, horizon length, scene complexity, and environmental uncertainty. 
TGPO consistently outperforms state-of-the-art prompting-based planners~\cite{ranasayplan_1,DELTA_11,LLM+P_1,Nl2plan_7} and existing reinforcement learning methods~\cite{GRPO,hu2501reinforce++}. 
It achieves particularly strong gains on complex, long-horizon tasks with abstract instructions and large-scale scenes, where implicit intent inference and compositional reasoning are critical. 
Moreover, TGPO generalizes effectively to novel tasks and environments, remains robust under stochastic dynamics, and integrates seamlessly with low-level executors, enabling scalable end-to-end robotic systems. Qualitative results are available on the project page.

Our contributions are threefold:
\textit{(1) Learning-based subgoal formulation.}
We cast long-horizon planning as learning a policy that generates verifiable subgoals, moving beyond prompting-based decomposition.
\textit{(2) Trace-guided policy optimization.}
We propose TGPO, a reinforcement learning algorithm that leverages external trace correction to improve structured generation under sparse and noisy rewards.
\textit{(3) Scalable long-horizon planning.}
We demonstrate that TGPO enables robust planning in large-scale environments with long horizons and abstract instructions, significantly outperforming existing methods.


\section{Related Work}\label{sec:related}

\textbf{LLMs for planning.}
Recent work leverages large language models (LLMs) for task planning by prompting them to generate action sequences or intermediate representations~\cite{ahn2022can, huang2022language_12, liang2023code, singh2022progprompt_6}. Approaches such as SayPlan~\cite{ranasayplan_1} and DELTA~\cite{DELTA_11} decompose long-horizon tasks into subproblems, often using structured inputs like scene graphs~\cite{3dsg_1,gu2024conceptgraphs}. While these methods demonstrate strong performance in simple settings, they rely on general-purpose LLMs at inference time and typically suffer from limited scalability as task ambiguity and constraint complexity increase.

\textbf{LLM--symbolic planning integration.}
Another line of work combines LLMs with symbolic planners~\cite{hu2023tree,zhao2023large,liu2022lang2ltl,guan2023leveraging,zhou2024isr}, typically using LLMs to generate PDDL~\cite{aeronautiques1998pddl} goals or problem specifications that are subsequently solved by off-the-shelf planners~\cite{fast_downward} (e.g., LLM+P~\cite{LLM+P_1}, NL2Plan~\cite{Nl2plan_7}). These methods benefit from the verifiability and optimality of symbolic planning, but depend on the LLM's ability to produce correct symbolic formulations~\cite{valmeekam2023planbench}. As a result, performance is often bottlenecked by the quality of prompting and does not directly improve with experience. In contrast, our approach treats subgoal generation as a learnable problem and optimizes it through reinforcement learning.

\textbf{Reinforcement learning for LLM.}
Reinforcement learning has been widely used to align and improve LLMs via methods such as RLHF~\cite{ouyang2022training}, GRPO~\cite{GRPO}, and Reinforce++~\cite{hu2501reinforce++}. These approaches typically rely on outcome-based reward signals and optimize generation at the token level. Consequently, they face significant challenges in abstract tasks characterized by sparse, ambiguous, and noisy rewards~\cite{cobbe2021training, lightman2023let}. In particular, credit assignment over intermediate reasoning steps remains difficult. Our method addresses this by incorporating external feedback through trace-level correction.

\textbf{Process supervision and external guidance.}
Recent work explores supervising intermediate reasoning processes, such as chain-of-thought supervision~\cite{wei2022chain,lightman2023let,yao2022react}, self-refinement~\cite{madaan2023self,huang-etal-2023-large}, and reflection-based~\cite{shinn2023reflexion,renze2024self} methods. These approaches improve reasoning quality by modifying intermediate outputs, but are usually applied at inference time or through supervised learning. In contrast, we integrate trace correction directly into reinforcement learning, enabling the policy to learn from corrected reasoning trajectories during optimization.

\textbf{Summary.}
In contrast to prior work that relies on prompting or post-hoc refinement, we formulate subgoal generation as a learnable policy (\secref{sec:problem}) and introduce a trace-guided reinforcement learning framework (\secref{sec:tgpo}) for optimizing structured reasoning under verifiable constraints.

\section{Problem Formulation}\label{sec:problem}
We consider long-horizon planning from natural language under structured constraints. Given a potentially abstract language instruction $q$, a structured representation of a large environment $\mathcal{S}$, e.g., a scene graph~\cite{gu2024conceptgraphs, armeni20193d}, and a PDDL domain $\mathcal{D}$ that symbolically specifies the robot's capabilities, the goal is to produce an action plan $A=\{a_k\}_{k=1}^K$ that satisfies the user's intent while remaining feasible under $\mathcal{D}$.

\textit{Structured subgoal generation.}
Instead of directly generating action sequences or a single PDDL goal, we formulate planning as generating a sequence of intermediate PDDL subgoals:
\begin{equation}
(T, G) = \pi_\theta(q, \mathcal{S}),
\end{equation}
where $T = \{t_k\}_{k=1}^K$ is a sequence of natural-language decomposition steps, and $G = \{g_k\}_{k=1}^K$ is a sequence of grounded symbolic subgoals. Each subgoal $g_k$ is expressed in a structured form (i.e., PDDL) and is associated with a subset of relevant objects.

\textit{Solver-grounded execution.}
Each subgoal $g_k$ defines a subproblem that can be solved by a symbolic planner:
\begin{equation}
a_k = \textsc{Solve}(\mathcal{D}, \mathcal{S}_k, g_k),
\end{equation}
where $\mathcal{D}$ denotes the planning domain and $\mathcal{S}_k$ is the updated environment state after executing previous subgoals. The final plan $A=\{a_k\}_{k=1}^K$ is obtained by composing the subplans.

\textit{Planning objectives.}
We define a reward function based on two criteria: (i) \emph{feasibility}, which checks whether each subgoal is solvable by the planner, and (ii) \emph{task completion}, which evaluates whether the composed plan satisfies the instruction. Formally, the reward for an output $(T, G)$ is:
\begin{equation}
R(T, G) = R_{\text{feas}}(G) \cdot R_{\text{comp}}(T, G),
\end{equation}
where $R_{\text{feas}} \in \{0,1\}$ indicates whether all subgoals are solvable, and $R_{\text{comp}} \in [0,1]$ measures task completion.

\textit{The learning challenge.}
Optimizing $\pi_\theta$ under this objective is challenging for two reasons. First, the reward is sparse and noisy, especially for abstract tasks where goal completion is difficult to verify. Second, errors often arise from intermediate reasoning steps (e.g., missing, overlapping or misordered subgoals), making credit assignment over the sequence difficult. These challenges motivate the trace-guided optimization strategy introduced in \secref{sec:tgpo}.

\begin{figure*}[!t] 
    \centering
    \includegraphics[width=\linewidth]{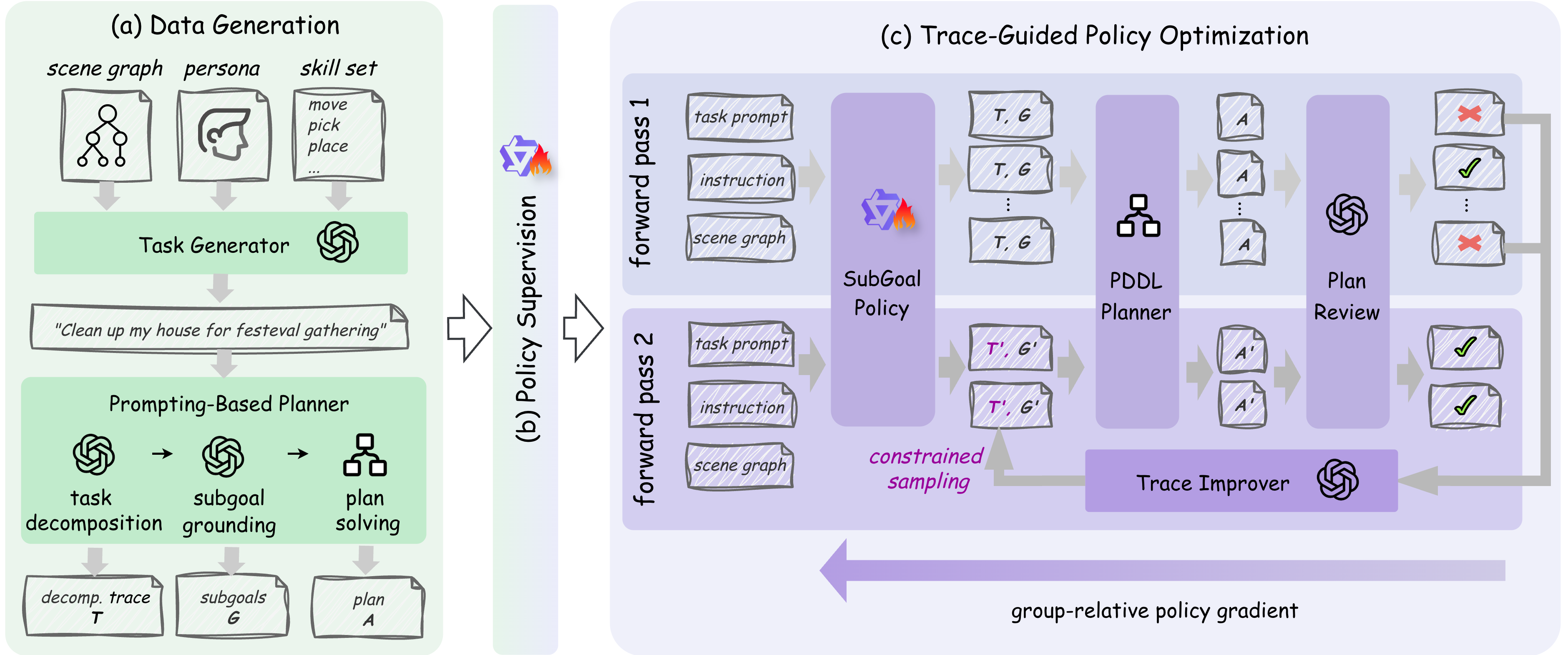} 
    \caption{
Overview of TGPO. 
\textit{(a) Data Generation}: Construction of a long-horizon planning dataset with task decomposition traces and grounded subgoals. 
\textit{(b) Policy Initialization}: Supervised fine-tuning (SFT) to initialize a structured generation policy. 
\textit{(c) Trace-Guided Policy Optimization}: A reinforcement learning framework that improves structured generation by incorporating external trace correction and constrained sampling, enabling effective learning under sparse and noisy rewards.
}
    \label{fig:overview}
\end{figure*}

\section{Trace-Guided Policy Optimization}\label{sec:tgpo}

We propose \textbf{Trace-Guided Policy Optimization (TGPO)}, a reinforcement learning algorithm for structured subgoal generation under sparse and noisy rewards. TGPO builds upon group-relative policy optimization (GRPO)~\cite{GRPO}, but differs in how learning signals are constructed and utilized: it introduces a two-pass rollout with external trace correction, and performs policy optimization over both free and trace-constrained trajectories, enabling structured credit assignment over intermediate reasoning steps.


Let $\pi_\theta$ denote the policy that generates structured outputs $(T, G)$ given instruction $q$ and environment $\mathcal{S}$. Following GRPO, we sample a group of $N$ candidate outputs $\{o_n\}_{n=1}^N$ and compute group-relative advantages based on their rewards $R(o_n)$.
The standard GRPO objective is:
\begin{equation} \label{eqn:grpo-objective}
\mathcal{L}_{\text{GRPO}}(\theta) =
\mathbb{E} \left[
\frac{1}{N} \sum_{n=1}^N
\min \big( \rho_n A_n, \text{clip}(\rho_n, 1-\epsilon, 1+\epsilon) A_n \big)
\right],
\end{equation}
where $\rho_n = \frac{\pi_\theta(o_n)}{\pi_{\theta_{\text{old}}}(o_n)}$ and $A_n$ is the group-relative advantage.
While effective, this formulation treats each output as a whole and relies solely on scalar rewards, making it difficult to correct errors arising from intermediate reasoning steps. 

TGPO augments GRPO with a second rollout pass that explicitly corrects intermediate reasoning traces (\figref{fig:overview}(c)).


\textbf{Free rollouts.}
We sample a group of $N$ trajectories from the current policy $\pi_\theta$:
\begin{equation}
o_n = (T_n, G_n) \sim \pi_\theta(\cdot \mid q, \mathcal{S}), \quad n = 1, \dots, N,
\end{equation}
where each output $o_n$ consists of a decomposition trace $T_n = \{t_{n,k}\}_{k=1}^{K_n}$ and a sequence of grounded subgoals $G_n = \{g_{n,k}\}_{k=1}^{K_n}$. 
The policy $\pi_\theta$ defines an autoregressive distribution over tokens that jointly generate $(T_n, G_n)$. Each trajectory is evaluated using the verifier-grounded reward $R(o_n)$.

\textbf{Trace correction.}
We identify a subset of low-quality trajectories based on their rewards:
\begin{equation}
\mathcal{I}_{\text{fail}} = \{ n \mid R(o_n) < \tau \},
\end{equation}
where $\tau$ is a predefined threshold.
For each $n \in \mathcal{I}_{\text{fail}}$, we apply a trace correction operator $\mathcal{C}$ that revises the decomposition trace:
\begin{equation}
\tilde{T}_n = \mathcal{C}(T_n; q, \mathcal{S}),
\end{equation}
where $\mathcal{C}$ is implemented by an auxiliary model that conditions on the instruction and environment, details are provided in Appendix~\ref{nips_app:training_details}.
The corrected trace $\tilde{T}_n = \{\tilde{t}_{n,k}\}_{k=1}^{\tilde{K}_n}$ is intended to address common failure modes in structured reasoning, including missing subtasks, incorrect temporal ordering, and incomplete grounding of task intent. Importantly, this step modifies only the intermediate reasoning trace, while leaving subgoal generation to the policy in the subsequent rollout.

\textbf{Constrained rollouts.}
For each corrected trace $\tilde{T}_n$, we perform a second rollout by conditioning the policy on $\tilde{T}_n$ and generating the corresponding subgoals:
\begin{equation}
\tilde{o}_n = (\tilde{T}_n, \tilde{G}_n) \sim \pi_\theta(\cdot \mid q, \mathcal{S}; \tilde{T}_n).
\end{equation}
Concretely, we implement this as token-level constrained sampling: tokens corresponding to the trace $\tilde{T}_n$ are \emph{forced} during decoding, while the remaining tokens (i.e., subgoals) are generated autoregressively. The resulting trajectories $\{\tilde{o}_n\}$ are then evaluated using the reward $R(\tilde{o}_n)$.


\textbf{Policy optimization.}
We combine trajectories from both rollout passes into an augmented group:
\begin{equation}
\mathcal{G} = \{o_n\}_{n=1}^N \cup \{\tilde{o}_n\}_{n \in \mathcal{I}_{\text{fail}}},
\end{equation}
where $\{o_n\}$ are trajectories from the free rollout and $\{\tilde{o}_n\}$ are generated via constrained rollout with corrected traces.
We compute group-relative advantages over $\mathcal{G}$ and apply a similar clipped objective as in \eqref{eqn:grpo-objective}:
\begin{equation}
\mathcal{L}_{\text{TGPO}}(\theta) =
\mathbb{E} \left[
\frac{1}{|\mathcal{G}|} \sum_{o \in \mathcal{G}}
\min \big( \rho(o) A(o), \text{clip}(\rho(o), 1-\epsilon, 1+\epsilon) A(o) \big)
\right],
\end{equation}
where $\rho(o) = \frac{\pi_\theta(o)}{\pi_{\theta_{\text{old}}}(o)}$.

Note that, during constrained rollout, TGPO computes the log-probability of the full sequence, so that policy gradients are computed with respect to all tokens in the sequence, including the constrained trace. 
Moreover, policy gradients in TGPO are computed over both original and trace-corrected trajectories. This effectively incorporates external corrections into the optimization process.
Note that, while constrained sampling introduces a deviation from the policy’s native exploration distribution, we do not explicitly correct for this mismatch via importance weighting. Instead, the bias is naturally mitigated in practice because constraints are applied only to the subset of failed trajectories and are restricted to intermediate reasoning tokens, leaving the remaining generation process on-policy. 


\section{Application to Large-Scale Household Planning}\label{sec:application}
We instantiate the proposed framework for long-horizon household planning by grounding the general formulation in Section~\ref{sec:problem} in mobile manipulation tasks in large indoor environments.
Each task is specified by a natural language instruction $q$ and a structured environment scene graph $\mathcal{S} = (V, E, X)$.
Here, $V$ denotes entities such as rooms, furniture, and objects; $E$ encodes spatial and semantic relations (e.g., \texttt{in}, \texttt{on}); and $X$ specifies symbolic state attributes (e.g., open/closed, clean/dirty). 
Further details on the scene graph representation are provided in Appendix~\ref{nips_app:sg_representation}.
Planning is performed over a household PDDL domain $\mathcal{D}$ that defines a wide set of predicates and operator schemas for mobile manipulation, details are provided in Appendix~\ref{nips_app:pddl_domain}. 
Given $(q, \mathcal{S})$, the objective is to generate a sequence of grounded subgoals that are valid under $\mathcal{D}$ and can be composed into an executable mobile manipulation plan via a symbolic planner.


\subsection{Training Data Generation}\label{sec:data-generation}
We construct a large-scale dataset by jointly sampling household environments and tasks with controlled diversity. \figref{fig:overview}(a) visualizes the data generation pipeline.

\textit{(a) Scene distribution.}
We construct the scene distribution from the Habitat Synthetic Scene Dataset (HSSD)~\cite{hssd}, which contains high-quality 3D indoor environments with realistic layouts and a diverse set of real-world objects. We directly reuse the textual scene graphs provided by HSSD.
To increase diversity while preserving realism, we further augment these base scene graphs by injecting additional objects and attributes conditioned on task requirements. This augmentation maintains plausible spatial and semantic consistency, resulting in a large set of scene graphs that reflect realistic environments while supporting diverse planning scenarios.

\textit{(b) Task distribution.}
As illustrated in \figref{fig:overview}(a), given a base scene $\mathcal{S}$ and a skill set derived from the planning domain $\mathcal{D}$, we synthesize task instructions using a large language model jointly conditioned on $\mathcal{S}$, $\mathcal{D}$, and a sampled user \textit{persona}. Each persona specifies attributes such as occupation, preferences, and situational context, inducing diverse task intents grounded in the same environment.
This persona-conditioned generation enables controlled diversity in task distribution, producing instructions that vary not only in surface form but also in underlying intent and abstraction. In particular, we explicitly vary the level of abstraction, ranging from explicit commands to high-level intent descriptions that require implicit reasoning and grounding. 
Further details regarding the synthetic dataset are provided in Appendix~\ref{nips_app:synthetic_dataset}.

\textit{(c) Annotation and filtering.}
For each $(q, \mathcal{S})$, we generate structured outputs $(T, G)$—including decomposition traces and grounded subgoals—using a prompting-based planner. This planner employs a general-purpose LLM (e.g., GPT~\cite{jaech2024openai}) to produce outputs in the same format as the TGPO policy (detailed in Appendix~\ref{app:planning_details}), but without task-specific training.
We retain only samples that satisfy both feasibility (all subgoals are solvable) and completion (the resulting plan fulfills the task intent), yielding high-quality supervision for learning.

\subsection{Training Pipeline}\label{sec:training-pipeline}
We adopt a two-stage training pipeline consisting of supervised fine-tuning (SFT, \figref{fig:overview}(b)) followed by reinforcement learning with TGPO (\figref{fig:overview}(c)).
In the first stage, we initialize the policy via supervised fine-tuning (SFT) on the constructed dataset using Qwen2.5-7B~\cite{qwen2.5} as the base model, enabling the policy to learn fundamental capabilities for task decomposition and subgoal grounding.
In the second stage, we apply TGPO to further optimize the policy via trace-guided reinforcement, refining intermediate reasoning, increasing task completion, and enforcing subgoal feasibility.
Further details regarding the training are provided in Appendix~\ref{nips_app:training_details}.

\section{Experiments}\label{sec:experiments}


\paragraph{Benchmarks.}
We evaluate our method on four complementary benchmarks.

\textit{(a) In-distribution synthetic tasks.}
We evaluate on novel in-distribution (ID) tasks drawn from the same scene and task generation processes as the training data. 
Importantly, while the training set includes only tasks that are solvable by a prompting-based planner, the test set is not subject to this filtering, and therefore is a broader and potentially more challenging set.
The test tasks are partitioned into three categories:
\begin{itemize}[leftmargin=20pt]
    \item \textbf{Easy} (146 tasks / 109 scenes): instructions explicitly specify the required \emph{action} and the \emph{objects of interest}, leading to short-horizon plans with few constraints. 
    Example: {``I'm in the bathroom, can you bring me a towel?''}.
    \item \textbf{Complex} (94 tasks / 75 scenes): explicit instructions with long-horizon (15.6 steps on average), constraint-heavy planning, typically involving multiple entities (\eg, objects, devices, and furniture) and cross-area coordination.
    Example: {``Retrieve the portable induction stove and hot-pot saucepan from the kitchen cabinet, place them on the kitchen island, fill the saucepan with water, and turn on the stove.''}.
    \item \textbf{Abstract} (108 tasks / 45 scenes): instructions that express user needs at an abstract level (6.4 words on average) without explicitly specifying the target objects or intermediate subgoals, requiring the model to infer complex and implicit intent; the planning horizon is comparable to the Complex subset (20 steps on average).
    Example: {``Clean up the house for a festival gathering.''}.
\end{itemize}

\textit{(b) Human tasks.}
We evaluate on human-authored instructions collected on scenes generated from the same pipeline. Participants are presented with visualized scene graphs and asked to propose tasks. After manual filtering for quality, the resulting test set consists of 50 tasks.
Human instructions are highly concise—averaging only 5.7 words—comprising a mix of explicit short-horizon tasks and abstract long-horizon tasks. See details in Appendix~\ref{nips_app:human_benchmark}.

\textit{(c) PARTNR~\cite{chang2024partnr}.}
We evaluate on the PARTNR benchmark, a public household task planning benchmark. Despite similar planning horizons (15.8 steps on average), PARTNR tasks are explicitly instructed. Scenes are also significantly smaller (65 nodes on average vs.\ 150 in our setting), including only task-relevant objects.

\textit{(d) BEHAVIOR-1K~\cite{li2023behavior}.}
We further evaluate on BEHAVIOR-1K, a public benchmark for household task planning and skill learning. We use the 50 household tasks from the 2025 BEHAVIOR Challenge.
These tasks feature long planning horizons (24.7 steps on average), but with explicit instructions (40.6 words on average) and substantially smaller scene graphs (11.5 nodes on average).

\textbf{Evaluation protocol and metrics.}
In \secref{sec:prompting-comparison}-\ref{sec:stochastic}, we evaluate high-level planning independently. In \secref{sec:system-level}, we further provide a system-level evaluation with low-level execution.

Given a policy output $(T, G)$, we evaluate the following two aspects of the high-level plan:

\textit{(a) Feasibility.}
We assess whether subgoals are executable under the given environment. For synthetic, human, and BEHAVIOR-1K tasks, feasibility is verified using PDDL solvers~\cite{fast_downward}, which check whether all subgoals are solvable. For PARTNR, we instead rely on the simulator to determine feasibility. This defines a feasibility score $S_{\text{feas}}(o_i) \in \{0,1\}$. $S_{\text{feas}}(o_i)=1$ means the entire plan is feasible.

\textit{(b) Task completion.}
We evaluate whether the composed plan satisfies the instruction. For synthetic and human tasks, we use an \textit{LLM-as-judge} protocol, LLM-Rubric~\cite{hashemi2024llm}. We query a large language model multiple times from different evaluation perspectives, including consistency between the final state and the instruction, alignment of decomposed subtasks with the instruction, action-sequence validity, and symbolic state-transition consistency. We then aggregate the scores via a weighted average. Details are provided in Appendix~\ref{nips_app:rubric_llm_evaluation}. For PARTNR and BEHAVIOR-1K, task completion is determined using their simulator's explicit goal, details are provided in Appendix~\ref{sec:benchmark_details}. This defines a task completion score $S_{\text{comp}}(o_i) \in [0,1]$.

Combining the above derives the \textit{Success Rate}, written as
$\textrm{SR}(o_i) = S_{\text{feas}}(o_i) \cdot S_{\text{comp}}(o_i)$. We also report the planning time $T$ when applicable.

\textbf{Baselines.}
We compare TGPO with three categories of baselines:

\textit{(a) Prompting-based planners.}
SayPlan~\cite{ranasayplan_1} performs task decomposition and action grounding in natural language, and iteratively re-plans via reflection.
DELTA~\cite{DELTA_11} formulates a holistic PDDL problem, and decomposes it into subproblems for efficient solving.
LLM+P and NL2Plan generate a PDDL problem and directly solve it.
We also include vanilla {Gemini-3-Pro} and {GPT-5} as baselines\footnote{We use \texttt{Gemini-3-Pro} (low reasoning setting) and \texttt{GPT-5} (no extra reasoning effort).}, prompting them to directly generate action plans assisted by generic chain-of-thought (CoT)~\cite{wei2022chain}. 

\textit{(b) Learning-based planners.}
SFT~\cite{ouyang2022training} learns subgoal generation using supervised learning. 
GRPO~\cite{GRPO} applies group-relative policy optimization to learn subgoal generation via RL.
Reinforce++~\cite{hu2501reinforce++} is an RL method based on REINFORCE, with reduced variance. 
\ourgrpo (E2E) is an ablated version of our method, learning to directly output action sequences.
All methods are trained with the same data and optimization settings for fair comparison.

\subsection{Performance Comparison with Prompting-Based Planners}\label{sec:prompting-comparison}

\begin{table*}[!t]
\centering
\small
\caption{Comparison with prompting-based baselines. Numbers are success rates (\%). 
$T$ denotes the average planning time per task.}
\label{tab:main_results}

\resizebox{\textwidth}{!}{
\begin{tabular}{lccccccc}
\toprule
& & \multicolumn{3}{c}{\textbf{In-Distribution}} & \multicolumn{3}{c}{\textbf{Out-of-Distribution}} \\
\cmidrule(lr){3-5} \cmidrule(lr){6-8}
\textbf{Method} & \textbf{T (s)} & Easy & Complex & Abstract & Human & PARTNR & BEHAVIOR-1K \\
\midrule

\gptfive{}      
& 8.0
& 38.3 $\pm$ 3.4  & 24.6 $\pm$ 1.0  & 9.1 $\pm$ 3.2 
& 21.7 $\pm$ 2.7  & 69.2 $\pm$ 1.2  & 35.1 $\pm$ 1.0 \\

\geminithreepro{}    
& 23.5
& 30.0 $\pm$ 2.3  & 16.8 $\pm$ 1.4  & 19.9 $\pm$ 2.9 
& 13.0 $\pm$ 3.0  & 71.8 $\pm$ 1.4  & 26.0 $\pm$ 1.1 \\

\midrule

\algsayplan{}          
& 63.7
& 47.5 $\pm$ 2.8  & 33.5 $\pm$ 2.2 & 22.4 $\pm$ 2.1 
& 26.5 $\pm$ 3.4 & 72.2 $\pm$ 1.8  & 43.8 $\pm$ 0.7 \\

\algdelta{}            
& 37.8
& 52.7 $\pm$ 2.6  & 22.0 $\pm$ 2.8 &  3.9 $\pm$ 1.3 
& 26.5 $\pm$ 2.1  & 69.3 $\pm$ 1.1 & 39.3 $\pm$ 1.5 \\

LLM+P
& 12.3
& 12.6 $\pm$ 1.0 &   2.1 $\pm$ 0.4 &   4.4 $\pm$ 0.8 
&  3.9 $\pm$ 1.8 & 67.2 $\pm$ 1.7 & 22.5 $\pm$ 1.2 \\

NL2Plan
& 18.9
& 12.8 $\pm$ 0.9 & 10.6 $\pm$  0.7 &  1.7 $\pm$ 0.6
&  4.5 $\pm$ 2.0 & 66.8 $\pm$ 1.0 & 21.0 $\pm$ 1.2 \\

\midrule

\ourgrpo      
& 3.1
& \textbf{88.1} $\pm$ 0.4 
& \textbf{77.3} $\pm$ 0.4 
& \textbf{70.2} $\pm$ 1.8 
& \textbf{73.6} $\pm$ 1.9  
& \textbf{82.6} $\pm$ 0.5  
& \textbf{70.3} $\pm$ 0.8  \\

\bottomrule
\end{tabular}
}
\vspace{-0.5cm}
\end{table*}
Table~\ref{tab:main_results} summarizes performance across all benchmarks. TGPO consistently outperforms prompting-based methods by a large margin, with gains that systematically increase as tasks become more long-horizon, abstract, and ambiguous. This trend highlights the advantage of our learning-based approach over prompting-based methods, particularly in handling implicit intent and maintaining feasibility over long-horizon tasks.

\textit{(a) In-distribution tasks.}
TGPO significantly outperforms prompting-based methods across all in-distribution settings, with improvements over the strongest baseline increasing from 67\% on easy tasks to 131\% on complex tasks and 213\% on abstract tasks, indicating that the advantage of TGPO becomes more pronounced as task complexity and ambiguity increase.

\textit{(b) Human tasks.}
The performance of \ourgrpo on human tasks is consistent with that on in-distribution complex and abstract tasks, achieving a 178\% improvement over the strongest prompting-based planner, demonstrating robust generalization to real-world task distributions.


\textit{(c) PARTNR.}
On PARTNR, where both tasks and environments are out-of-distribution, TGPO achieves a success rate of 82.6\%, demonstrating strong generalization. Since PARTNR tasks are explicitly specified and involve simple constraints and small scene graphs, they are easier for all methods.

\textit{(d) BEHAVIOR-1K.}
TGPO also successfully generalizes to BEHAVIOR-1K, where both environments and tasks differ significantly from the training distribution, achieving a success rate of 70.3\%. Compared to the strongest baseline, this corresponds to a 61\% improvement. The substantial performance gain highlights the robustness of TGPO under severe distribution shift.

\subsection{Performance Comparison with Learning-Based Planners}\label{sec:learning-comparison}
\begin{table*}[t]
\centering
\small
\setlength{\tabcolsep}{6pt}
\caption{Comparison with learning-based baselines. Numbers are success rates (\%).}
\label{tab:learning_comparison}

\resizebox{\textwidth}{!}{
\begin{tabular}{lcccccc}
\toprule
& \multicolumn{3}{c}{\textbf{In-Distribution}} & \multicolumn{3}{c}{\textbf{Out-of-Distribution}} \\
\cmidrule(lr){2-4} \cmidrule(lr){5-7}
\textbf{Method} & Easy & Complex & Abstract & Human & PARTNR & BEHAVIOR-1K \\
\midrule

Qwen2.5-7B (Base) 
& 0.0 & 0.0 & 0.0 
& 0.0 & 27.8 $\pm$ 1.0 & 12.5 $\pm$ 0.7 \\

SFT         
& 83.2 $\pm$ 0.5 & 71.1 $\pm$ 0.6 & 38.1 $\pm$ 1.7 
& 55.8 $\pm$ 2.2 & 71.7 $\pm$ 1.3 & 63.7 $\pm$ 1.1\\

GRPO        
& 84.1 $\pm$ 0.8 & 76.0 $\pm$ 0.2 & 51.1 $\pm$ 2.4 
& 65.9 $\pm$ 2.6 & 76.4 $\pm$ 0.9 & 67.1 $\pm$ 0.8 \\

Reinforce++ 
& 83.7 $\pm$ 0.6 & 77.1 $\pm$ 0.3 & 54.0 $\pm$ 1.5 
& 65.5 $\pm$ 3.3 & 75.2 $\pm$ 1.2 & 68.3 $\pm$ 1.3\\

\midrule

TGPO (E2E)          
& 72.4 $\pm$ 0.5 & 55.4 $\pm$ 0.2 & 44.1 $\pm$ 2.0 
& 54.1 $\pm$ 2.2 & 62.3 $\pm$ 0.6 & 54.4 $\pm$ 1.0\\

TGPO (Subgoal, ours)       
& \textbf{88.1} $\pm$ 0.4 
& \textbf{77.3} $\pm$ 0.4 
& \textbf{70.2} $\pm$ 1.8 
& \textbf{73.6} $\pm$ 1.9  
& \textbf{82.6} $\pm$ 0.5  
& \textbf{70.3} $\pm$ 0.8 \\

\bottomrule
\end{tabular}
}
\end{table*}

Table~\ref{tab:learning_comparison} compares TGPO with learning-based baselines, also serving as an ablation study of different training strategies.
We also provide a scalability test of \ourgrpo in Appendix~\ref{nips_app:scalability}.

\textit{(a) Base model.}
The Qwen2.5-7B base model completely fails on synthetic and human tasks, and achieves marginal success on PARTNR and BEHAVIOR-1K, due to their simplified scene graphs. These results reaffirm that mid-scale open models cannot solve large-scale planning tasks without domain-specific adaptation.

\textit{(b) TGPO vs.\ SFT.}
Compared to supervised fine-tuning (SFT), TGPO achieves substantially higher performance, especially on complex and abstract tasks. This suggests that reinforcement learning provides a significantly higher performance upper bound by optimizing directly for plan feasibility under environment constraints and task completion under abstract instructions.

\textit{(c) TGPO vs.\ GRPO / Reinforce++.}
The key difference between TGPO and standard RL methods (GRPO, Reinforce++) lies in the use of trace revision as structured guidance during training. Without trace revision, models perform comparably on simple tasks but degrade rapidly as task complexity and abstraction increase. In contrast, TGPO remains robust.
This behavior can be explained by the difficulty of credit assignment under sparse and noisy rewards. For long-horizon tasks, successful trajectories are rare, making exploration inefficient. Trace revision helps the policy quickly discover high-quality trajectories. For abstract tasks, reward signals become ambiguous and noisy; trace revision provides additional reinforcement signals that stabilize policy learning.

\textit{(d) TGPO (Subgoal) vs.\ TGPO (E2E).}
Comparing TGPO (Subgoal) with TGPO (E2E) highlights the importance of structured outputs and symbolic planning. The end-to-end variant must simultaneously interpret the task, decompose it, and generate a feasible action sequence, which significantly increases the learning burden. Failures often arise from violating constraints in the action sequence, leading to infeasible plans.
In contrast, TGPO (Subgoal) separates problem formulation from execution: the policy generates structured subgoals, while a symbolic planner tackles constraint satisfaction. This decomposition enables efficient learning and substantially improves performance on complex tasks.

\subsection{Performance under Stochastic Environments and Execution Failures}\label{sec:stochastic}

Real-world robotic systems often operate under stochastic dynamics and execution uncertainty. Under such circumstances, the TGPO policy can be coupled with probabilistic PDDL formulations~\cite{younes2004ppddl1}. In fact, the generated subgoals can be applied to any domain sharing the same predicate and operator definitions. This allows replacing the deterministic domain with a probabilistic PDDL model at inference time, enabling closed-loop planning under stochastic action outcomes.
We evaluate this setting by integrating TGPO with online Monte Carlo Tree Search (MCTS)~\cite{kocsis2006bandit} in a stochastic environment with a 10\% per-action failure probability. Despite the presence of execution uncertainty, the system achieves 80.5\% success on easy tasks, 70.8\% on complex tasks, and 62.3\% on abstract tasks, validating its effectiveness under stochastic dynamics.

\subsection{System-Level Performance of Integrated Planning and Execution}
\label{sec:system-level}

We further evaluate the TGPO policy in a full robotic pipeline by integrating it with low-level executors.
We conduct experiments in the BEHAVIOR-1K 3D simulator using its household environments. We construct a set of 10 \textit{recombined tasks} by reordering and recombining previously demonstrated low-level skills into novel task sequences (see Appendix~\ref{nips_app:low_level_execution}). No demonstrations are provided for these tasks. This setup shifts the evaluation from imitating demonstrations (as in the BEHAVIOR-1K challenge) to planning and compositional generalization.

The robotic system is provided with textual scene graphs, online visual observations, and task instructions. The TGPO policy acts as a high-level planner that generates a sequence of high-level actions. Each action is executed using either parameterized skills (e.g., navigation) or a neural policy trained via imitation learning for complex primitives (e.g., pouring, opening). 
The neural policy~\cite{larchenko2025task} was trained on long-horizon demonstrations of the original tasks in BEHAVIOR-1K. In our system, however, it is restricted to local execution of primitive skills. 


Results show that the holistic neural policy generalizes poorly to recombined tasks, achieving only 23\% end-to-end task completion due to its reliance on demonstrated action plans. In contrast, our two-level system preserves strong high-level planning capability, with 90\% of generated plans passing feasibility verification, and achieves 36\% end-to-end task completion on the recombined tasks. For comparison, the end-to-end task completion rate on the original training tasks is 26\%. The substantial gap between high-level planning performance and end-to-end execution mainly reflects imperfections in the low-level VLA executor.
These results demonstrate that the TGPO policy enables compositional generalization by recombining a fixed set of low-level skills to solve novel long-horizon tasks, providing a principled approach to scaling complex robot behaviors beyond imitating demonstrations.
Execution demos and qualitative analysis of the recombined tasks are provided on the project website.

\section{Conclusion \& Limitations}\label{sec:conclusion_limitations}

We present \ourgrpo, a framework for long-horizon robot task planning that learns to generate verifiable subgoals from abstract natural language instructions. 
By combining a learned subgoal-generation policy with symbolic planners, \ourgrpo enables scalable planning in large environments with complex constraints. 
\ourgrpo directly optimizes subgoal feasibility, consistency, and task completion, while a trace-guided training scheme improves learning under sparse and noisy supervision. 
Experiments across diverse benchmarks demonstrate that \ourgrpo significantly outperforms prior prompting-based planners and learning-based methods, with particularly strong gains on complex, long-horizon, and abstract tasks.

Despite these results, several limitations remain. 
First, \ourgrpo relies on structured scene representations, which may lose information compared to raw visual observations; extending the framework to operate directly on visual inputs is an important direction. 
Second, the method assumes a predefined, comprehensive symbolic domain; learning or adapting domains across tasks and environments is a key direction for future work. 
Third, our current formulation assumes full observability, leaving open the integration of perception uncertainty and active exploration in real-world settings.

\bibliographystyle{unsrt}
\bibliography{references}

\clearpage
\appendix 
\counterwithin{figure}{section}
\counterwithin{table}{section}
\counterwithin{equation}{section}

\section{Planning Pipeline Details}\label{app:planning_details}

\begin{figure}[h] 
    \centering
    \includegraphics[width=0.9\textwidth]{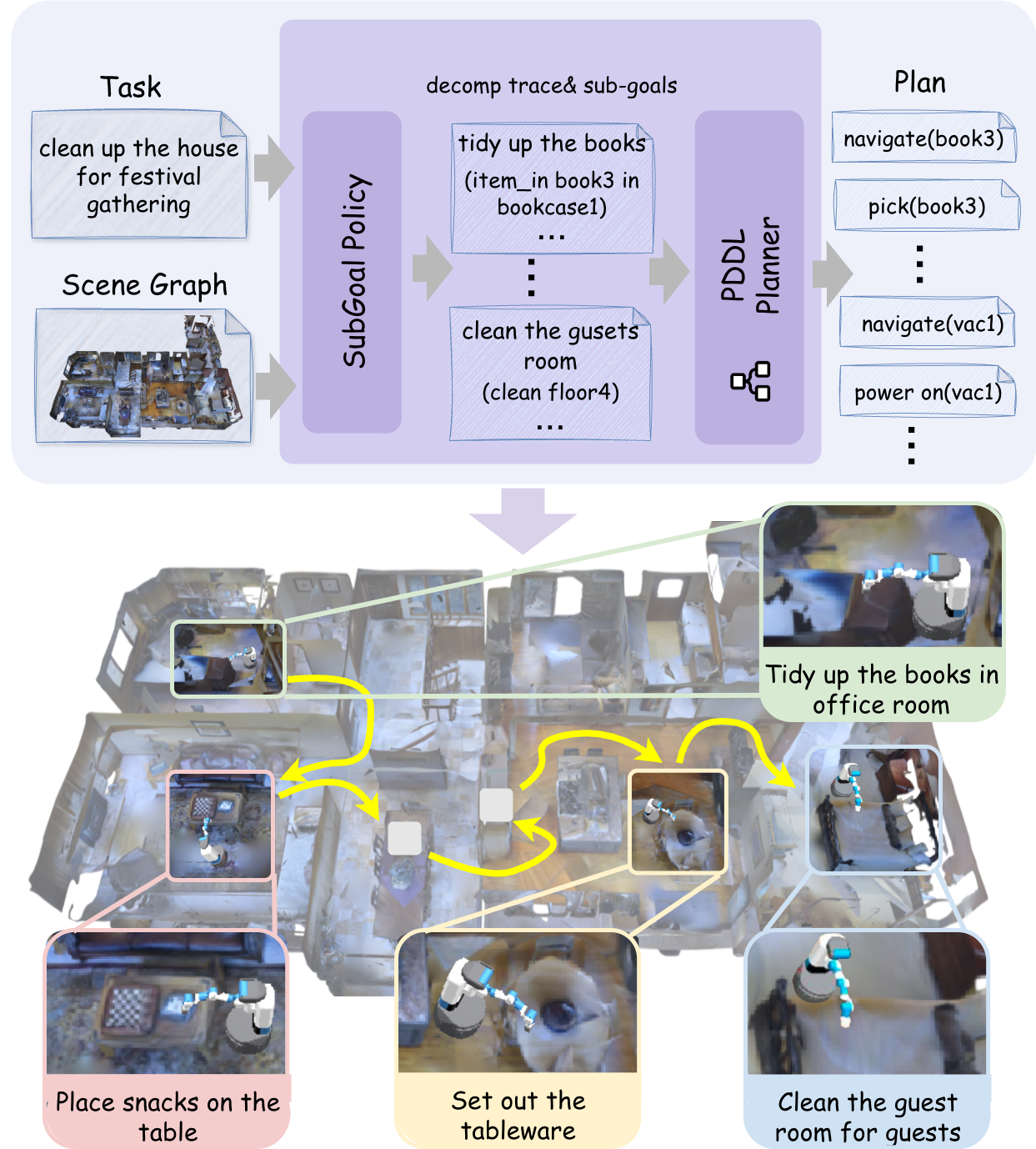} 
    \caption{In large-scale environments, the subgoal policy receives abstract instructions and a scene graph, and generates a decomposition trace with corresponding subgoals. These subgoals are then solved using a PDDL planner, resulting in an executable long-horizon plan that satisfies the user's requirements.}
    
    \label{fig:ahat-plan}
\end{figure}

This appendix provides comprehensive implementation details of the planning pipeline, which is illustrated in Figure~\ref{fig:ahat-plan}. The pipeline operates as follows: taking a user instruction and an environment scene graph (detailed in Appendix~\ref{nips_app:sg_representation}) as inputs, the subgoal policy systematically breaks down the instruction by first generating task decomposition steps (Appendix~\ref{app:trace_format}), followed by a corresponding sequence of grounded subgoals (Appendix~\ref{app:subgoal_format}). Subsequently, the pipeline processes these subgoals incrementally: for each subgoal, a rule-based $\text{Construct}(\cdot)$ procedure (Appendix~\ref{app:construct_rules}) translates the current scene graph into a specific PDDL problem file, which is then solved by a PDDL solver (Appendix~\ref{app:planner_config}) to generate a subplan. Finally, all resulting subplans are concatenated to yield the complete executable plan. The underlying PDDL domain file is provided in Appendix~\ref{nips_app:pddl_domain}.

\subsection{Scene Graph Representation and State Attributes}\label{nips_app:sg_representation}

We formalize the environment as a scene graph (SG) comprising typed nodes and semantic edges. Specifically, the nodes represent entities categorized as \texttt{room}, \texttt{furniture}, \texttt{object}, \texttt{agent}, or \texttt{person}. The edges capture spatial relations—namely, containment (\texttt{in}) and placement (\texttt{on}). To facilitate planning and action feasibility checking, each entity is augmented with a set of state attributes, as detailed in Table~\ref{tab:sg_states}. Finally, to interface with the language model, this structured SG is serialized into a textual format. A complete example of this serialized representation is provided in Listing~\ref{lst:sg_serialization_example}.

\begin{table*}[h]
\centering
\small
\setlength{\tabcolsep}{6pt}
\caption{State attributes stored in the scene graph for planning.}
\label{tab:sg_states}
\begin{tabular}{p{0.18\linewidth} p{0.78\linewidth}}
\toprule
\textbf{Entity type} & \textbf{State attributes} \\
\midrule
Furniture / Objects &
\texttt{is\_open}, \texttt{can\_be\_opened} \newline
\texttt{is\_powered\_on}, \texttt{is\_powerable} \newline
\texttt{is\_clean}, \texttt{is\_cleaning\_tool},\newline
\texttt{require\_floor\_cleaner}, \texttt{requires\_water\_to\_clean} \newline
\texttt{has\_faucet}, \texttt{is\_filled} \newline
\texttt{is\_heating\_device}, \texttt{is\_heater}, \texttt{is\_heated}
\\
\midrule
Person &
\texttt{agent\_at}, \texttt{p\_holding}
\\
\midrule
Agent &
\texttt{agent\_at}, \texttt{holding}, \texttt{handempty}
\\
\bottomrule
\end{tabular}
\end{table*}

\begin{lstlisting}[
    caption={Example of serialized textual scene graph (SG).},
    label={lst:sg_serialization_example},
    basicstyle=\ttfamily\small,
    breaklines=true,
    frame=single,
    backgroundcolor=\color{gray!5},
    columns=fullflexible,
    postbreak=\mbox{\textcolor{red}{$\hookrightarrow$}\space}
]
{
  "rooms": [
    {
      "name": "bathroom_1",
      "states": {
        "is_clean": false,
        "require_floor_cleaner": true
      }
    },
    {
      "name": "bedroom_1",
      "states": {
        "is_clean": false,
        "require_floor_cleaner": true
      }
    }
  ],
  "furnitures": [
    {
      "states": {
        "is_clean": false
      },
      "name": "bed_4",
      "type": "bed"
    }
  ],
  "objects": [
    {
      "states": {
        "requires_water_to_clean": true
      },
      "name": "bottle_10",
      "type": "bottle"
    },
    {
      "states": {
        "requires_water_to_clean": true
      },
      "name": "bowl_11",
      "type": "bowl"
    }
  ],
  "agent": [
    {
      "name": "agent_0",
      "location": "corridor_19"
    }
  ],
  "person": [
    {
      "name": "person_0",
      "location": "bathroom_4"
    }
  ],
  "links": [
    ["bedroom_1", "bed_4", "in"]
  ]
}
\end{lstlisting}

\subsection{Decomposition Trace Output Format}\label{app:trace_format}

We output a \emph{decomposition trace} as an ordered list of natural-language (or structured)
subtasks. Below we show a representative example.

\begin{lstlisting}[
    caption={Task instruction prompt used for decomposition (example).},
    label={app:task_prompt_example},
    basicstyle=\ttfamily\small,
    breaklines=true,
    frame=single,
    backgroundcolor=\color{gray!5},
    columns=fullflexible,
    postbreak=\mbox{\textcolor{red}{$\hookrightarrow$}\space}
]
Task: "Create a classy dinner atmosphere and tidy afterward"
\end{lstlisting}

\begin{lstlisting}[
    caption={Decomposition trace format (example).},
    label={app:decomposition_trace_example},
    basicstyle=\ttfamily\small,
    breaklines=true,
    frame=single,
    backgroundcolor=\color{gray!5},
    columns=fullflexible,
    postbreak=\mbox{\textcolor{red}{$\hookrightarrow$}\space}
]
### Task Decomposition ###
1. Pick_up_tablecloth_20_from_dining_table_18
2. Place_tablecloth_20_on_dining_table_18
3. Pick_up_centerpiece_84_from_dining_table_18
4. Place_centerpiece_84_next_to_tablecloth_20_on_dining_table_18
5. Pick_up_candle_141_from_dining_table_18
6. Place_candle_141_next_to_centerpiece_84_on_dining_table_18
7. Pick_up_salt_shaker_58_from_dining_table_18
8. Place_salt_shaker_58_next_to_candle_141_on_dining_table_18
9. Pick_up_pepper_grinder_59_from_dining_table_18
10. Place_pepper_grinder_59_next_to_salt_shaker_58_on_dining_table_18
11. Pick_up_candle_holder_60_from_dining_table_18
12. Place_candle_holder_60_next_to_pepper_grinder_59_on_dining_table_18
13. Pick_up_napkin_holder_22_from_dining_table_20
14. Place_napkin_holder_22_next_to_candle_holder_60_on_dining_table_18
15. Pick_up_wine_bottle_142_from_dining_table_20
16. Place_wine_bottle_142_next_to_napkin_holder_22_on_dining_table_18
\end{lstlisting}

\subsection{PDDL Subgoal and Relevant Object Set Format}\label{app:subgoal_format}

For each subtask $t_k$, the policy outputs (i) a set of grounded PDDL subgoals, and
(ii) an implicitly related object set. We keep the output in a structured textual format:

\begin{lstlisting}[
    caption={Subgoal grounding output format (example).},
    label={app:subgoal_grounding_example},
    basicstyle=\ttfamily\small,
    breaklines=true,
    frame=single,
    backgroundcolor=\color{gray!5},
    columns=fullflexible,
    postbreak=\mbox{\textcolor{red}{$\hookrightarrow$}\space}
]
### Subgoal Grounding ###
1. Pick_up_tablecloth_20_from_dining_table_18
   Subtask goal: ['(holding agent_0 tablecloth_20)']
   Implicitly related object: []
2. Place_tablecloth_20_on_dining_table_18
   Subtask goal: ['(item_on_surface tablecloth_20 dining_table_18)']
   Implicitly related object: []
3. Pick_up_centerpiece_84_from_dining_table_18
   Subtask goal: ['(holding agent_0 centerpiece_84)']
   Implicitly related object: []
4. Place_centerpiece_84_next_to_tablecloth_20_on_dining_table_18
   Subtask goal: ['(next_to centerpiece_84 tablecloth_20 dining_table_18)']
   Implicitly related object: []
5. Pick_up_candle_141_from_dining_table_18
   Subtask goal: ['(holding agent_0 candle_141)']
   Implicitly related object: []
6. Place_candle_141_next_to_centerpiece_84_on_dining_table_18
   Subtask goal: ['(next_to candle_141 centerpiece_84 dining_table_18)']
   Implicitly related object: []
...
\end{lstlisting}

\subsection{Problem file construction procedure}\label{app:construct_rules}

We formulate a subtask-specific PDDL problem instance by grounding the predicted subgoal set $g_k$ and its related objects $obj_k$ within the current scene graph state $SG_k$.

Given $(SG_k, g_k, obj_k)$, the construction follows these steps:
\begin{enumerate}[leftmargin=1.2em]
  \item \textbf{Extract core objects.} Parse all object symbols appearing in the grounded PDDL goals $g_k$ and take their union with the implicitly related object set $obj_k$.
  \item \textbf{Expand with parent nodes.} For each core object, retrieve its parent nodes within the SG hierarchy (e.g., the containing furniture and room) based on \texttt{in} and \texttt{on} spatial links.
  \item \textbf{Extract initial state.} From $SG_k$, extract the current predicates of all collected entities, capturing spatial relations (placement/containment) and relevant state attributes.
  \item \textbf{Assemble the PDDL problem.} Define the object universe using the extracted entities, instantiate the initial state with the retrieved predicates, and set the goal condition to $g_k$.
\end{enumerate}

\subsection{Subgoal Planner Details}\label{app:planner_config}

To compute the final action sequences, we employ the off-the-shelf \texttt{Fast Downward} planner configured with the \texttt{lama-first} alias. To strictly preserve causal dependencies among subgoals, we formulate the planning process incrementally rather than solving for all goals simultaneously. Specifically, if a subtask $k$ comprises a sequence of subgoals $g_k=\{A,B,C\}$ alongside a related object set $O$, we solve them sequentially as follows:
\begin{enumerate}[leftmargin=1.2em]
  \item Formulate and solve a problem with goal $A$ and its required objects (derived from $A$ and $O$) to obtain $\texttt{subplan}_{k,A}$; execute this subplan to update $SG$.
  \item Formulate a subsequent problem with the cumulative goal $(A \wedge B)$ and objects required for both goals to obtain $\texttt{subplan}_{k,B}$; execute and update $SG$.
  \item Formulate the final problem with goal $(A \wedge B \wedge C)$ and all corresponding objects to obtain $\texttt{subplan}_{k,C}$.
  \item Concatenate the solutions to form the complete subtask plan: $\texttt{subplan}_{k}=\texttt{subplan}_{k,A}\oplus \texttt{subplan}_{k,B}\oplus \texttt{subplan}_{k,C}$.
\end{enumerate}

\section{Training Details}
\label{nips_app:training_details}

\subsection{SFT Details}\label{app:sft_details}
We initialize subgoal policy from a Qwen2.5-7B base model and perform supervised fine-tuning (SFT) as a warm-up stage before launching TGPO. The complete set of hyperparameters is summarized in Table~\ref{tab:sft_hparams}.

\begin{table}[h]
\centering
\caption{SFT hyperparameters.}
\label{tab:sft_hparams}
\small
\setlength{\tabcolsep}{8pt}
\begin{tabular}{l c}
\toprule
\textbf{Hyperparameter} & \textbf{Value} \\
\midrule
Base model & Qwen2.5-7B \\
Batch size & 32 \\
SFT epochs & 3 \\
Cutoff length (\texttt{cutoff\_len}) & 15000 \\
Learning rate & $5\times 10^{-6}$ \\
Warm-up ratio & 0.1 \\
\bottomrule
\end{tabular}
\end{table}

\paragraph{Data format.}
Each training instance follows the subgoal policy output schema: given an instruction $q$ and a textual scene graph $SG$, the target output is a structured completion containing (i) a decomposition trace $T$ and (ii) grounded subgoals $\{(g_k,\mathrm{obj}_k)\}_{k=1}^{K}$.

\subsection{GRPO and TGPO Hyperparameters}\label{app:tgpo_hparams}

In this section, we detail the hyperparameter configurations used for training both the baseline Group Relative Policy Optimization (GRPO) and our proposed Trace-Guided Policy Optimization (TGPO). To ensure a fair and rigorous comparison, we maintain identical settings for all shared training hyperparameters across both methods, only tuning the components specific to TGPO. The complete set of hyperparameters is summarized in Table~\ref{tab:tgpo_hparams}.

\begin{table}[h]
\centering
\caption{TGPO and GRPO hyperparameters.}
\label{tab:tgpo_hparams}
\small
\setlength{\tabcolsep}{8pt}
\begin{tabular}{l c}
\toprule
\textbf{Hyperparameter} & \textbf{Value} \\
\midrule
Learning rate & $5\times 10^{-6}$ \\
Batch size & 128 \\
Mini-batch size & 32 \\
Rollout number & 8 \\
$\texttt{TGPO\_fix\_num\_per\_rollout}$ & 2 \\
\bottomrule
\end{tabular}
\end{table}

\subsection{Task Completion Review Details}\label{app:completion_reward_prompt}
Task completion review is performed using Qwen3-235B-A with a single-query evaluation protocol for training efficiency. We first convert the solved plan into a detailed natural-language description, and then evaluate whether the resulting execution satisfies the task objective. In addition to the generated plan summary, the evaluator is provided with the task instruction and scene information. The model then determines whether the plan successfully completes the task. Prompt details are provided on the project page: \textit{(\href{https://tgpo2026.github.io/TGPO/appendix_materials/prompts/training_prompts/Task_completion_reviewer_prompt.md}{View Prompt})}.

\subsection{Trace Improver Details}\label{app:trace_improver_protocol}
We use Qwen3-235B-A22B-Thinking-2507 as the auxiliary model for trace refinement. During trace refinement, the model is provided with the task instruction, scene information, task-completion reviewer feedback, and the original reasoning trace. The auxiliary model revises the reasoning trace to address errors in task decomposition, temporal ordering, and intent grounding, and outputs a corrected trace for the subsequent constrained rollout stage. Prompt details are provided on the project page: \textit{(\href{https://tgpo2026.github.io/TGPO/appendix_materials/prompts/training_prompts/Trace_Improver_prompt.md}{View Prompt})}.

\subsection{Training Compute Resources}\label{app:training_compute_resources}
All experiments are conducted on 8×H200 GPUs. Supervised fine-tuning is performed for approximately 48 hours, followed by 100 hours of reinforcement learning with TGPO.

\section{Rubric LLM Evaluation Details}\label{nips_app:rubric_llm_evaluation}

We adopt a multi-dimensional LLM-as-judge protocol to evaluate whether a composed plan satisfies a given instruction. We decompose the evaluation into four orthogonal dimensions, query each independently with differentiated inputs, and aggregate the scores via a weighted average. This design mitigates the variance inherent in single-call evaluation.

The pipeline operates in two stages: a deterministic state computation followed by a multi-perspective LLM evaluation using GPT-5 and Gemini-3-Pro.

\subsection{Deterministic State Computation via PDDL}
To prevent LLM hallucination when summarizing plan accomplishments, we deterministically compute the final state by applying each action's PDDL effects to the initial scene graph $\mathcal{S}_0$ using a forward simulator. For a plan $\pi = (a_1, \dots, a_n)$:
\begin{equation}
    \mathcal{S}_n = \text{apply}(a_n, \text{apply}(a_{n-1}, \dots \text{apply}(a_1, \mathcal{S}_0)))
\end{equation}
The final state $\mathcal{S}_n$ is extracted as a text-formatted set of grounded predicates (e.g., \texttt{item\_on\_surface(plate\_1, dining\_table)}).

\subsection{Multi-Perspective LLM Scoring}
We employ GPT-5 and Gemini-3-Pro as our evaluation LLMs, querying each model four times in parallel. Each query corresponds to a distinct evaluation dimension and receives a strictly differentiated subset of information, ensuring orthogonal assessment:

\begin{table}[htbp]
    \centering
    \caption{Multi-perspective evaluation dimensions ensuring orthogonal assessment.}
    \label{tab:evaluation_dimensions}
    \small
    \renewcommand{\arraystretch}{1.5}
    \begin{tabular}{@{} l c p{2cm} p{6.5cm} @{}}
        \toprule
        \textbf{Dimension} & \textbf{Weight} & \textbf{Input} & \textbf{Evaluation Focus \& Prompt Link} \\ 
        \midrule
        Goal Achievement & 0.4 & Final state \newline + Instruction & Judges purely by outcome: whether the execution satisfies the instruction goals. \textit{(\href{https://tgpo2026.github.io/TGPO/appendix_materials/prompts/rubric_llm_evaluation/Goal_Achievement_prompt.txt}{View Prompt})} \\
        
        Intent Understanding & 0.2 & Scene + Plan \newline + Instruction & Enables constraint-aware evaluation: selecting the best available option demonstrates correct intent. \textit{(\href{https://tgpo2026.github.io/TGPO/appendix_materials/prompts/rubric_llm_evaluation/Intent_Understanding_prompt.txt}{View Prompt})} \\
        
        Completeness & 0.2 & Plan \newline + Instruction & Evaluates subtask coverage at the goal level (e.g., ``set the table'' counts as one covered subtask). \textit{(\href{https://tgpo2026.github.io/TGPO/appendix_materials/prompts/rubric_llm_evaluation/Completeness_prompt.txt}{View Prompt})} \\
        
        Execution Quality & 0.2 & Plan \newline + Instruction & Checks for unreverted state changes, self-contradictions, navigation efficiency, and action count proportionality. \textit{(\href{https://tgpo2026.github.io/TGPO/appendix_materials/prompts/rubric_llm_evaluation/Execution_Quality_prompt.txt}{View Prompt})} \\
        \bottomrule
    \end{tabular}
\end{table}

\subsection{Score Aggregation}
Each dimension independently outputs a discrete score $d_{i,m} \in \{0.00, 0.25, 0.50, 0.75, 1.00\}$ with a rationale. This 5-point scale further reduces variance. For model $m \in \{\text{GPT-5}, \text{Gemini-3-Pro}\}$, the score is a weighted sum:
\begin{equation}
    s_m = \sum_{i=1}^{4} w_i \cdot d_{i,m}
\end{equation}
\textbf{Critical Failure Override:} If D1 identifies that a core instructed object is entirely wrong, or if the plan ignores the instruction, $s_m$ is capped at 0.2.

To ensure robustness, the final reported score is the average across both models:
\begin{equation}
    s_{\text{final}} = \frac{s_{\text{GPT-5}} + s_{\text{Gemini-3-Pro}}}{2}
\end{equation}

\section{Scalability Study}\label{nips_app:scalability}

\begin{figure}[t] 
    \centering 
    \includegraphics[width=0.95\textwidth]{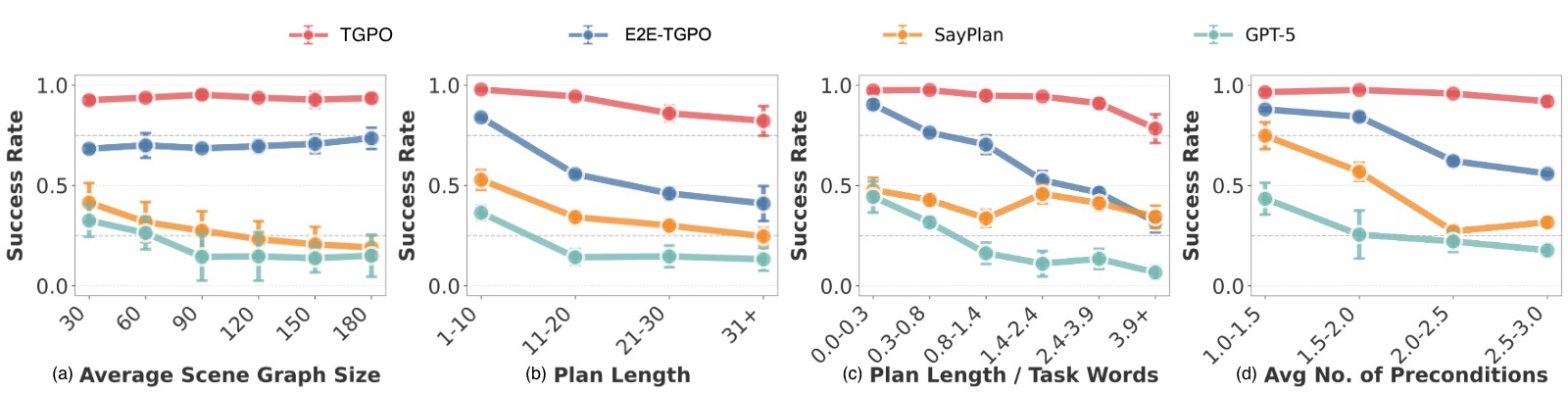} 
    \caption{Scalability Test Results showing the variation of success rate with: (a) Scene Graph Size, (b) Plan Length, (c) Task Abstractness, and (d) Constraint Complexity.} 
    \label{fig:scalability}
\end{figure}

We conduct scalability tests to assess how \ourgrpo scales with increasing task complexity. \figref{fig:scalability} shows that enlarging the environment, extending the planning horizon, increasing instruction abstractness, and increasing action-constraints complexity lead to only modest performance drops for \ourgrpo, indicating strong scalability and reasoning robustness. In contrast, baselines degrade substantially along one or more axes.

\paragraph{Scalability on Environment Size}
We measure scalability by the \emph{average scene-graph size} (30–180 nodes). As shown in \figref{fig:scalability}(a), \ourgrpo remains stable as graph size increases, with minimal degradation (16\%). In contrast, SayPlan and GPT-5 degrade significantly, with grounding errors exacerbated by larger graphs. \ourgrpo and E2E-\ourgrpo are both trained with RL on our constructed dataset, whose scene graphs span a wide range of scales (50--350 nodes; avg.\ $\sim$150), improving robustness to large environments at test time.

\paragraph{Scalability on Plan Length}
Horizon length is scaled by plan length (1–10, 11–20, 21–30, 31+ actions). As shown in Fig.~\ref{fig:scalability}(b), \ourgrpo's performance drops by only 16\% at the longest horizon, compared to 43\% for E2E-\ourgrpo. This resilience is due to \ourgrpo’s \emph{decompose-then-solve} structure, which prevents error accumulation by solving subplans sequentially, while E2E-\ourgrpo suffers from cascading errors in monolithic action sequences.

\paragraph{Scalability on Instruction Abstractness}
We quantify abstractness by the ratio of plan length to instruction word count. As abstractness increases (0–3.9+), \ourgrpo drops by 19\%, while E2E-\ourgrpo drops the most (58\%). This is due to the fact that in the end-to-end setup, after \ourgrpo trace improvement, the generated samples often fail to maintain feasibility and degrade into a GRPO-like behavior. As a result, E2E-\ourgrpo struggles to leverage \ourgrpo's benefits effectively, limiting its ability to handle abstract instructions.

\paragraph{Scalability on Constraint Complexity}
We measure constraint complexity by the average number of preconditions per action. As complexity increases (1.0–3.0), \ourgrpo's performance declines by just 3\%, while \algsayplan (42\%), E2E-\ourgrpo (32\%), and \gptfive (26\%) degrade significantly. \ourgrpo’s symbolic planner enforces preconditions and effects, maintaining feasibility even under dense constraints, unlike purely neural generation methods.

\section{Low-Level Execution Details}\label{nips_app:low_level_execution}

\subsection{VLA Execution}\label{app:vla_execution}

We adopt ~\cite{larchenko2025task}, which ranks first on the BEHAVIOR-1K leaderboard, as our low-level execution policy. During execution, we continuously extract environment state information from the simulator to determine high-level action termination. Specifically, the completion of each action is verified through environment state transitions.
High-level action execution is controlled by stage signals provided as inputs to the low-level execution policy, enabling structured guidance over action progression. 

\subsection{Motion Planning}\label{app:motion_planning}

Navigation is implemented using the A* algorithm to provide reliable long-distance motion planning. To bridge the gap between classical motion planning and low-level policy execution, we adopt a hybrid strategy: A* navigation terminates when the agent reaches within a 1.5-meter radius of the target location, after which fine-grained motion is completed by the VLA model.

During navigation execution, the robot arm joints are kept fixed to avoid unnecessary motion interference, ensuring stability and consistency between navigation and manipulation phases.

\begin{figure}[t]
    \centering
    \includegraphics[width=0.9\linewidth]{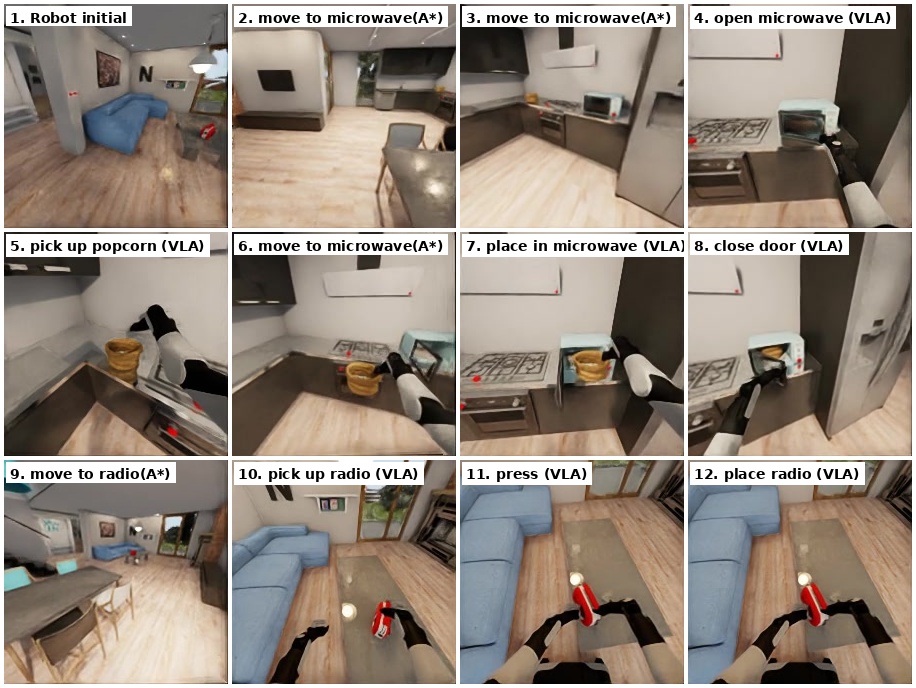}
    \caption{Visualization of the ``Preparing popcorn for a radio evening'' task.}
    \label{fig:combine_task}
\end{figure}

\subsection{Combined Tasks}\label{app:combined_tasks}

To evaluate compositional generalization, we construct combined tasks by merging partial or complete subtasks from existing tasks into new, semantically coherent household tasks. All non-navigation actions in these combined tasks are drawn from the training distribution, avoiding out-of-distribution execution by the VLA model.

The corresponding scenes are generated by merging the original scenes associated with the selected task components. This preserves environmental consistency while enabling more complex task structures. A full list of combined tasks is provided in Table~\ref{tab:combine_tasks}.
We visualize an example combined task in Fig.~\ref{fig:combine_task}. Complete videos are provided on the project page: \url{https://tgpo2026.github.io/TGPO/}

\begin{table*}[t]
\centering
\caption{Combined task descriptions and their corresponding source tasks.}
\label{tab:combine_tasks}

\small
\renewcommand{\arraystretch}{1.2}

\begin{tabularx}{\textwidth}{
|>{\raggedright\arraybackslash}p{3.2cm}
|>{\raggedright\arraybackslash}X
|>{\raggedright\arraybackslash}p{4.6cm}|}
\hline

\textbf{Combined Task} &
\textbf{Instruction} &
\textbf{Source Tasks} \\
\hline

Preparing popcorn for a radio evening
& Heat a popcorn bag in the microwave, then turn on the radio in the living room.
& make microwave popcorn \& turning on radio \\
\hline

Clean up kitchen after party
& Collect soda cans into the trash, move pizzas to the fridge, and put bowls into the sink.
& cleaning up plates and food \& picking up trash \\
\hline

Unload groceries and store food
& Bring the grocery sack from the car to the kitchen, then store items in the fridge and cabinets.
& carrying in groceries \& storing food \\
\hline

Tidy up children's room toys
& Put board games, puzzles, and a ball into the toy box, then place dice, teddy bears, and games into the bookcase.
& collecting children's toys \& picking up toys \\
\hline

Prep vegetables and cook
& Dice vegetables on the chopping board, then cook cabbage and chili in the frying pan.
& cook cabbage \& slicing vegetables \\
\hline

Evening relaxation setup
& Bring bottles from the fridge to the coffee table, then light a fire in the fireplace.
& bringing water \& setting the fire \\
\hline

Prepare bedroom for sleep
& Move the book to the nightstand and place sandals by the bed, then organize the desk.
& getting organized for work \& tidying bedroom \\
\hline

Quick lunch from fridge
& Cook hot dogs in the microwave, then cook bacon in the frying pan.
& cook bacon \& cook hot dogs \\
\hline

Seasonal decoration swap
& Store Halloween decorations in a cabinet, then set up Christmas decorations in the living room.
& putting away Halloween decorations \& putting up Christmas decorations inside \\
\hline

Sort vegetables and clear table
& Sort vegetables into mixing bowls, then pack leftovers into containers and refrigerate.
& clearing food from table into fridge \& sorting vegetables \\
\hline

\end{tabularx}
\end{table*}




\section{Benchmark Details and Analysis}\label{sec:benchmark_details}

\subsection{Behavior Benchmark Settings}\label{sec:behavior_benchmark}
\paragraph{Action Space Mapping}
The action names in our fixed PDDL domain differ from those used in the Behavior-1K benchmark. Therefore, we construct a mapping between our action space and the Behavior-1K action space to enable consistent execution and evaluation across the two systems.

\begin{table}[h]
\centering
\small
\caption{Action space mapping between our fixed PDDL domain and the Behavior-1K benchmark.}
\label{tab:action_mapping}
\begin{tabular}{ll}
\toprule
\textbf{Ours} & \textbf{Behavior-1K} \\
\midrule
navigate & NAVIGATE\_TO \\

pick\_from\_surface & GRASP \\
pick\_from\_receptacle & GRASP \\

place\_on\_surface & PLACE\_ONTOP \\
place\_in\_receptacle & PLACE\_INSIDE \\

open & OPEN \\
close & CLOSE \\

power\_on & TOGGLE\_ON \\
power\_off & TOGGLE\_OFF \\

pour\_into & pour \\
\bottomrule
\end{tabular}
\end{table}

\begin{figure}[t]
    \centering
    \includegraphics[width=0.9\linewidth]{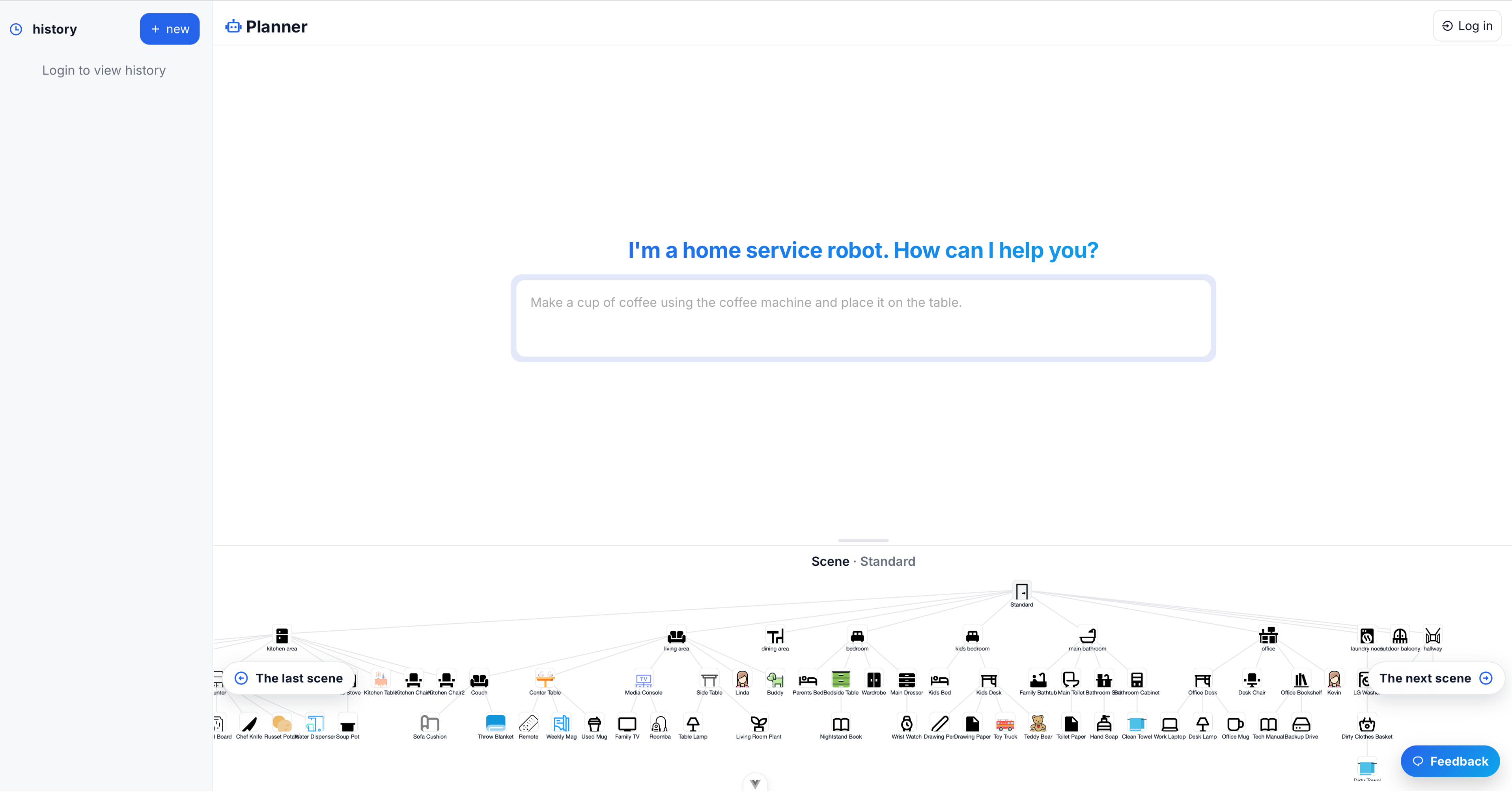}
    \caption{Screenshot of the web-based dialogue interface used to collect real human household task instructions. The interface visualizes the scene graph to provide environmental context, and participants are informed that their conversations may be used for research purposes.}
    \label{fig:human_web}
\end{figure}

\begin{table*}[tbp]
    \centering
    \small
    \setlength{\tabcolsep}{6pt}
    
    \caption{Easy task examples.}
    \label{tab:easy_task_examples}
    \begin{tabular}{l p{0.82\linewidth}}
    \toprule
    \textbf{ID} & \textbf{Instruction} \\
    \midrule
    E1  & I'm in the bathroom --- can you bring me a towel? \\
    E2  & Move a musical instrument from the living room to the bedroom. \\
    E3  & I want to do some crafts --- can you bring scissors, colored paper and glue to the couch? \\
    E4  & Could you fill my reusable water bottle and give it to the person on the stairs, please? \\
    \bottomrule
    \end{tabular}
    
    \vspace{1.5em} 
    
    \caption{Complex task examples.}
    \label{tab:complex_task_examples}
    \begin{tabular}{l p{0.82\linewidth}}
    \toprule
    \textbf{ID} & \textbf{Instruction} \\
    \midrule
    C1  & Refill the user's reusable water bottle and deliver it to them. \\
    C2  & Prepare a cup of boiled water and deliver it to the user. \\
    C3  & On a chilly winter morning, prepare the bathroom by pouring hot water from the kettle into a bowl placed on the bathroom counter, then bring a clean cup for rinsing. \\
    C4  & Go to the kitchen, get the mug from bench, fill it at sink, and bring the filled mug to person in bathroom and hand it over. \\
    C5  & Take a glass from the dining table, fill it with water at the kitchen sink, and hand the filled glass to person in empty\_room. \\
    \bottomrule
    \end{tabular}

    \vspace{1.5em} 
    
    \caption{Abstract task examples.}
    \label{tab:abstract_task_examples}
    \begin{tabular}{l p{0.82\linewidth}}
    \toprule
    \textbf{ID} & \textbf{Instruction} \\
    \midrule
    A1  & Create a classy dinner atmosphere and tidy afterward \\
    A2  & Prepare tomorrow's morning routine station. \\
    A3  & Get ready for a cozy movie marathon. \\
    A4  & Prepare a comforting breakfast and serve it \\
    A5  & Prepare the home for guests tonight. \\
    \bottomrule
    \end{tabular}
\end{table*}

\begin{table*}[tbp]
\centering
\caption{Human-authored task examples.}
\label{tab:human_task_examples}
\small
\setlength{\tabcolsep}{6pt}
\begin{tabular}{l p{0.55\linewidth} p{0.22\linewidth}}
\toprule
\textbf{ID} & \textbf{Instruction} & \textbf{Category} \\
\midrule
H1  & I'm thirsty. Give me a glass of water. & Explicit short-horizon \\
H2  & I'm kind of in the mood for coffee. & Abstract long-horizon \\
H3  & Give me some tea. & Explicit short-horizon \\
H4  & Mop the floor and clean it all up. & Explicit short-horizon \\
H5  & Prepare breakfast. & Abstract long-horizon \\
\bottomrule
\end{tabular}
\end{table*}

\paragraph{Evaluation Metrics and Success Criteria}
We validate plan feasibility using constraint rules extracted from the simulator for high-level planning. Task completion is evaluated based on the completion rate of the BDDL goals provided by the simulator.

\subsection{In-distribution Benchmark}
Examples from the in-distribution benchmark, including Easy, Complex, and Abstract tasks, are shown in Tables~\ref{tab:easy_task_examples}, \ref{tab:complex_task_examples}, and \ref{tab:abstract_task_examples}.

\subsection{Human Benchmark}\label{nips_app:human_benchmark}

We develop a web-based dialogue interface for a household service robot to collect \emph{real} human-authored household tasks.
The website visualizes the underlying scene graph to provide users with environmental context.
We publicly release the survey link and explicitly inform participants that their chat logs may be used for this research.
After legality and compliance screening, we obtain \textbf{50} real human household tasks.
A screenshot of the dialogue website is shown in Fig.~\ref{fig:human_web}.
Representative human task examples, together with their task categories, are shown in Table~\ref{tab:human_task_examples}.
The complete set of human-authored tasks is available on the project page:
\textit{(\href{https://tgpo2026.github.io/TGPO/appendix_materials}{View Tasks})}.

\section{Baseline implementation details}\label{app:baselines_prompts_budget}

This section details the configuration of the baseline models used in our experiments. Specifically, we provide the exact prompt templates deployed within the baseline pipelines, alongside the specific inference budgets allocated for evaluation.

\begin{table}[!h]
\centering
\caption{Baseline prompting pipelines and latency-matched inference budgets used in evaluation.}
\label{tab:baseline_budget}
\small
\setlength{\tabcolsep}{6pt}
\begin{tabular}{l c p{0.50\linewidth}}
\toprule
\textbf{Baseline} & \textbf{Avg. Time (s)} & \textbf{Budget / Setting} \\
\midrule
GPT-5   & 8.0  & No extra reasoning effort (latency-matched). \\
Gemini-3.0-pro & 23.5 & Low reasoning level (latency-matched). \\
SayPlan & 63.7 & Iterative prompting / replanning pipeline. \\
DELTA  & 37.8 & Decompose--formulate--solve per-subtask pipeline. \\
\bottomrule
\end{tabular}
\end{table}

\paragraph{General purpose LLM details.}
We directly prompt the general-purpose model to generate long-horizon plans under the same input interface (instruction + scene graph).The detailed semantic search prompt and the replan prompt is provided on project page: 
\textit{(\href{https://tgpo2026.github.io/TGPO/appendix_materials/prompts/baseline_prompts}{View Prompt})}

\paragraph{SayPlan prompt.}
We follow the original SayPlan-style pipeline prompt, adapted to our scene-graph interface. The detailed semantic search prompt and the replan prompt are provided on the project page:\textit{(\href{https://tgpo2026.github.io/TGPO/appendix_materials/prompts/baseline_prompts}{View Prompt})}.

\paragraph{DELTA details.}
DELTA comprises three prompting steps: (i) filter the scene graph to task-relevant subgraphs, (ii) decompose the task into subtasks, and (iii) generate a PDDL problem file for each subtask. For fair comparison, we directly use our fixed PDDL domain file to solve the problem, skipping the PDDL domain file generation process of DELTA. The detailed prompt is provided at the project page: \textit{(\href{https://tgpo2026.github.io/TGPO/appendix_materials/prompts/baseline_prompts}{View Prompt})}.

\paragraph{LLM+P details.}
We use GPT-5 as the LLM and follow the original LLM+P-style prompt, adapted to our setting. The low performance is mainly caused by poor-quality problem file generation in large environments and long-horizon planning tasks.
Prompt details are provided on the project page: \textit{(\href{https://tgpo2026.github.io/TGPO/appendix_materials/prompts/baseline_prompts}{View Prompt})}.

\paragraph{NL2Plan details}
For fair comparison, we skip NL2Plan's domain-file generation step and directly use our fixed PDDL domain file, while using GPT-5 as the underlying LLM. Prompt details are provided on the project page:
\textit{(\href{https://tgpo2026.github.io/TGPO/appendix_materials/prompts/baseline_prompts}{View Prompt})}.

\paragraph{End-to-end (E2E) training details.}
For the end-to-end variant, we prompt the model to first decompose the task into a tree-structured reasoning trace, and then ground the decomposed subtasks into executable actions. The full training-time prompt template is provided on the project page: \textit{(\href{https://tgpo2026.github.io/TGPO/appendix_materials/prompts/baseline_prompts}{View Prompt})}.

\section{Synthetic Dataset}\label{nips_app:synthetic_dataset}
We summarize key statistics and provide representative examples of the scene graphs, personas, and annotated human instruction planning tasks used to construct our dataset.

\begin{figure}[t]
    \centering
    \begin{subfigure}[t]{0.32\linewidth}
        \centering
        \includegraphics[width=\linewidth]{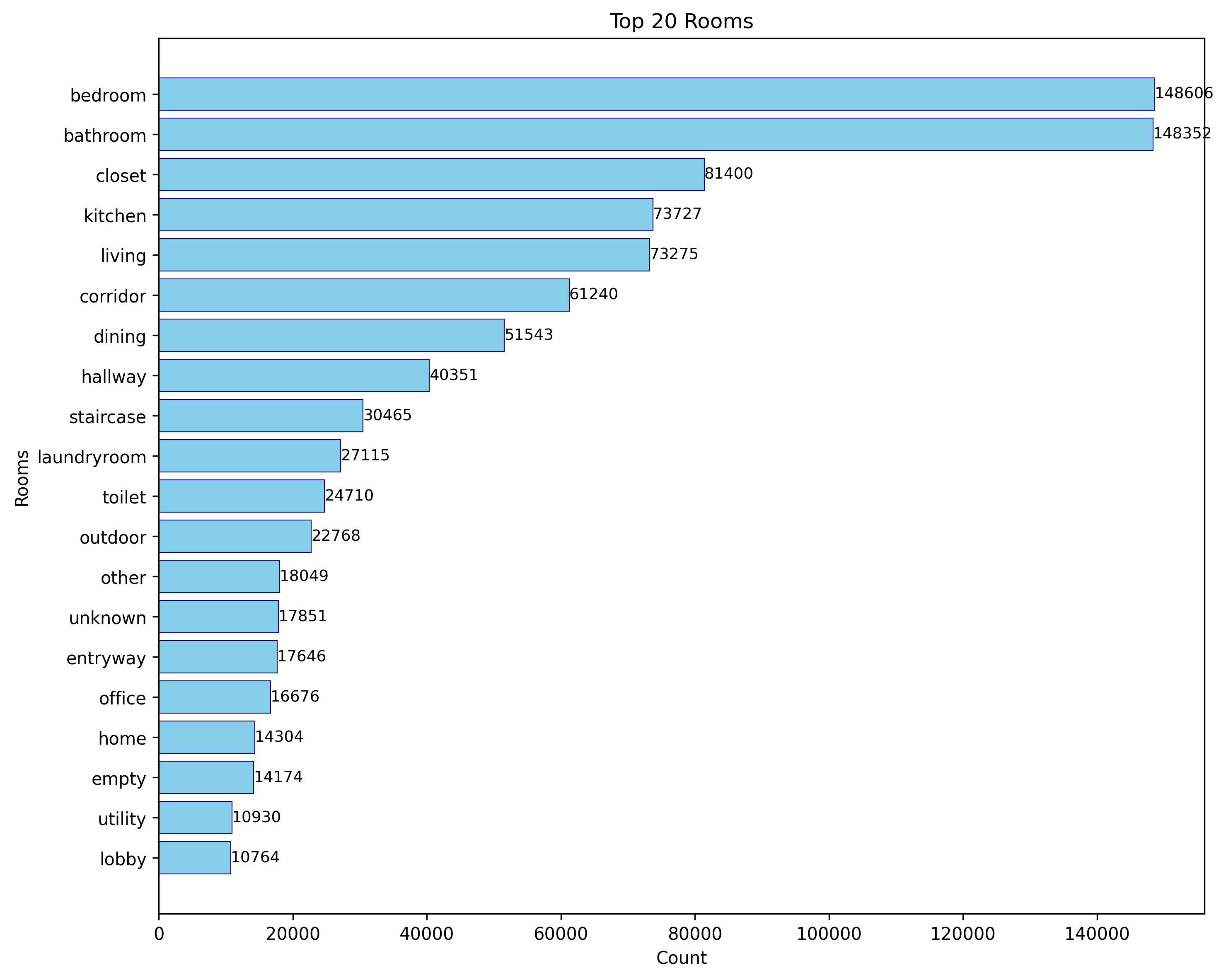}
        \caption{Top-20 rooms.}
        \label{fig:top20_rooms}
    \end{subfigure}\hfill
    \begin{subfigure}[t]{0.32\linewidth}
        \centering
        \includegraphics[width=\linewidth]{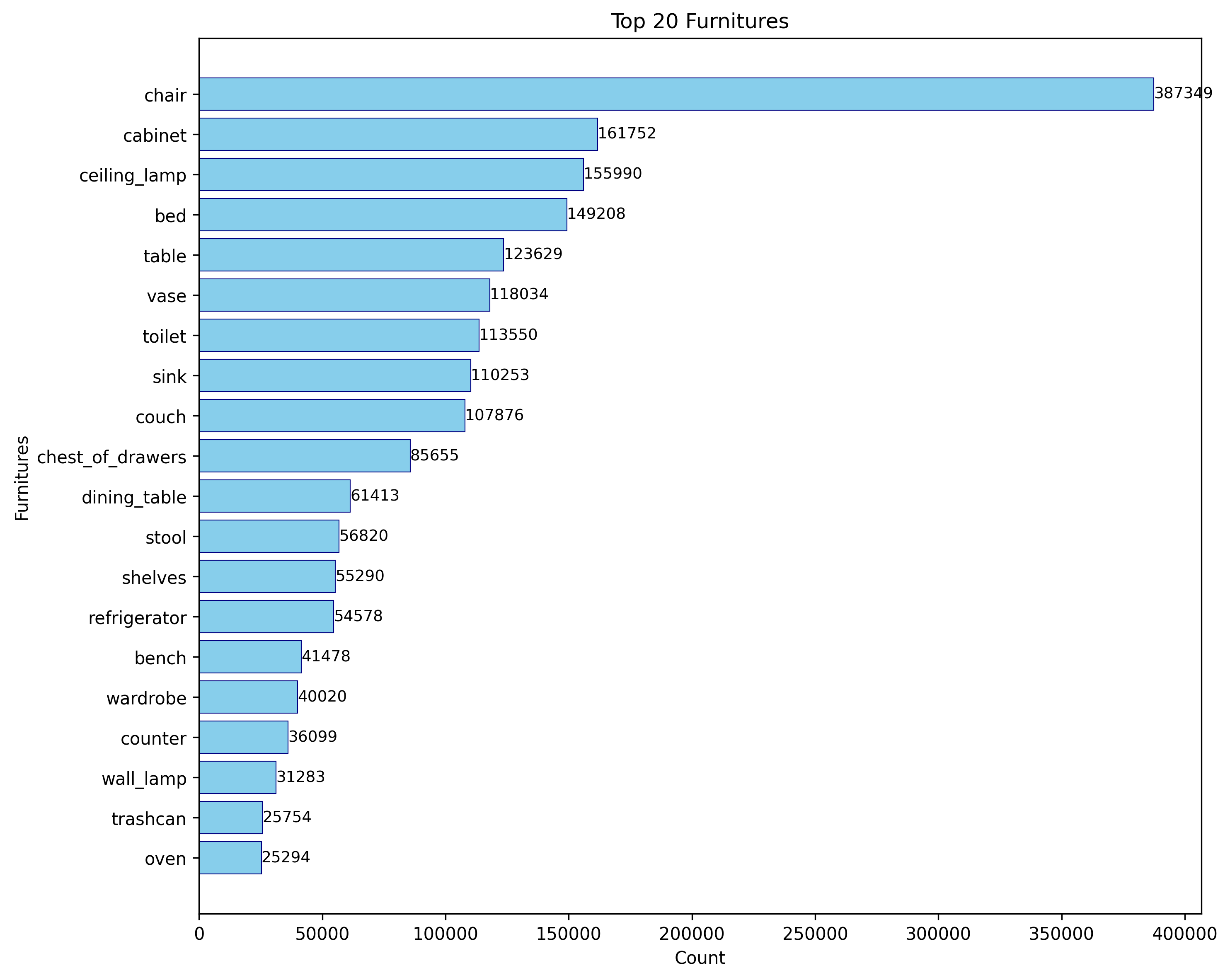}
        \caption{Top-20 furniture categories.}
        \label{fig:top20_furnitures}
    \end{subfigure}\hfill
    \begin{subfigure}[t]{0.32\linewidth}
        \centering
        \includegraphics[width=\linewidth]{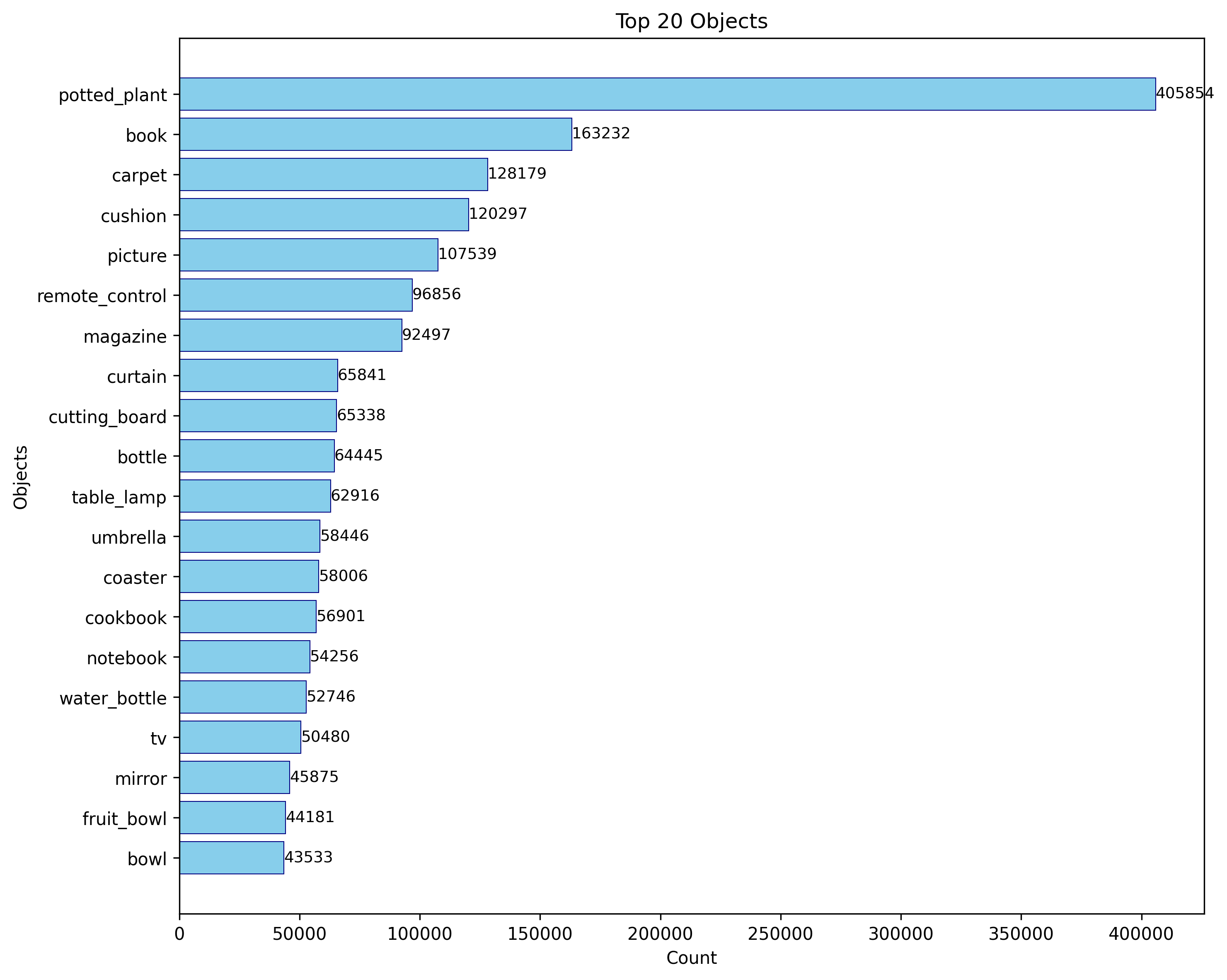}
        \caption{Top-20 object categories.}
        \label{fig:top20_objects}
    \end{subfigure}

    \caption{Entity frequency statistics over the \textbf{67.1k} enriched scene instances. We visualize the \textbf{top-20} most frequent rooms, furniture categories, and object categories, respectively.}
    \label{fig:scene_top20_stats}
\end{figure}

\paragraph{Scene Graphs}
We construct 67.1k enriched scene instances by augmenting 308 base scene graphs with task-relevant objects and relations.
Each enriched scene contains on average 142 nodes, reflecting diverse household layouts.
Across the 67.1k enriched scenes, we report the top-20 most frequent \textbf{rooms}, furniture categories, and object categories in Fig.~\ref{fig:scene_top20_stats}, highlighting common household compositions that the planner must reason about.

\begin{figure}[t]
    \centering

    \begin{subfigure}[t]{0.46\linewidth}
        \centering
        \includegraphics[width=\linewidth]{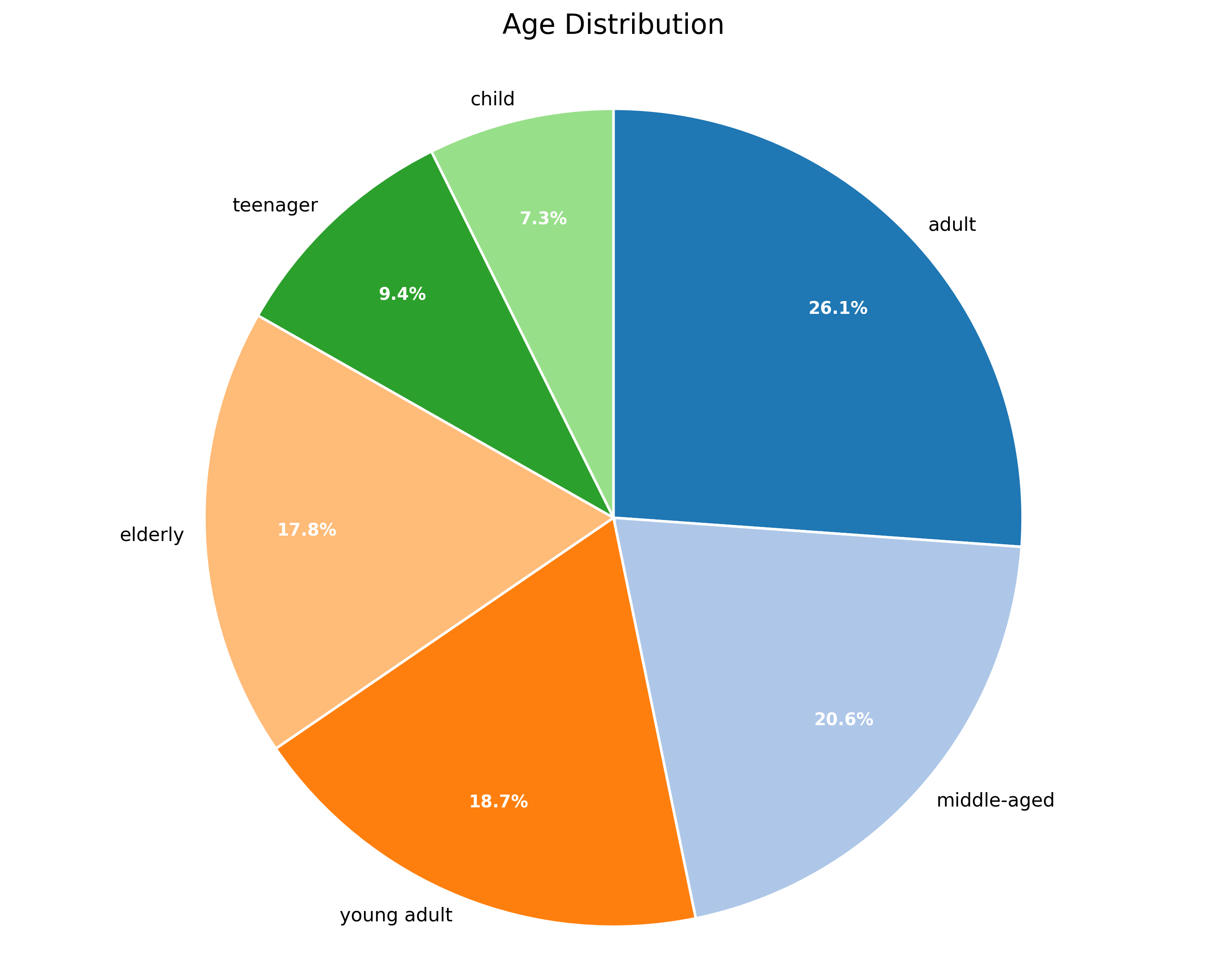}
        \caption{Age-group distribution of the \textbf{1677} synthesized personas.}
        \label{fig:persona_age_dist}
    \end{subfigure}
    \hfill
    \begin{subfigure}[t]{0.46\linewidth}
        \centering
        \includegraphics[width=\linewidth]{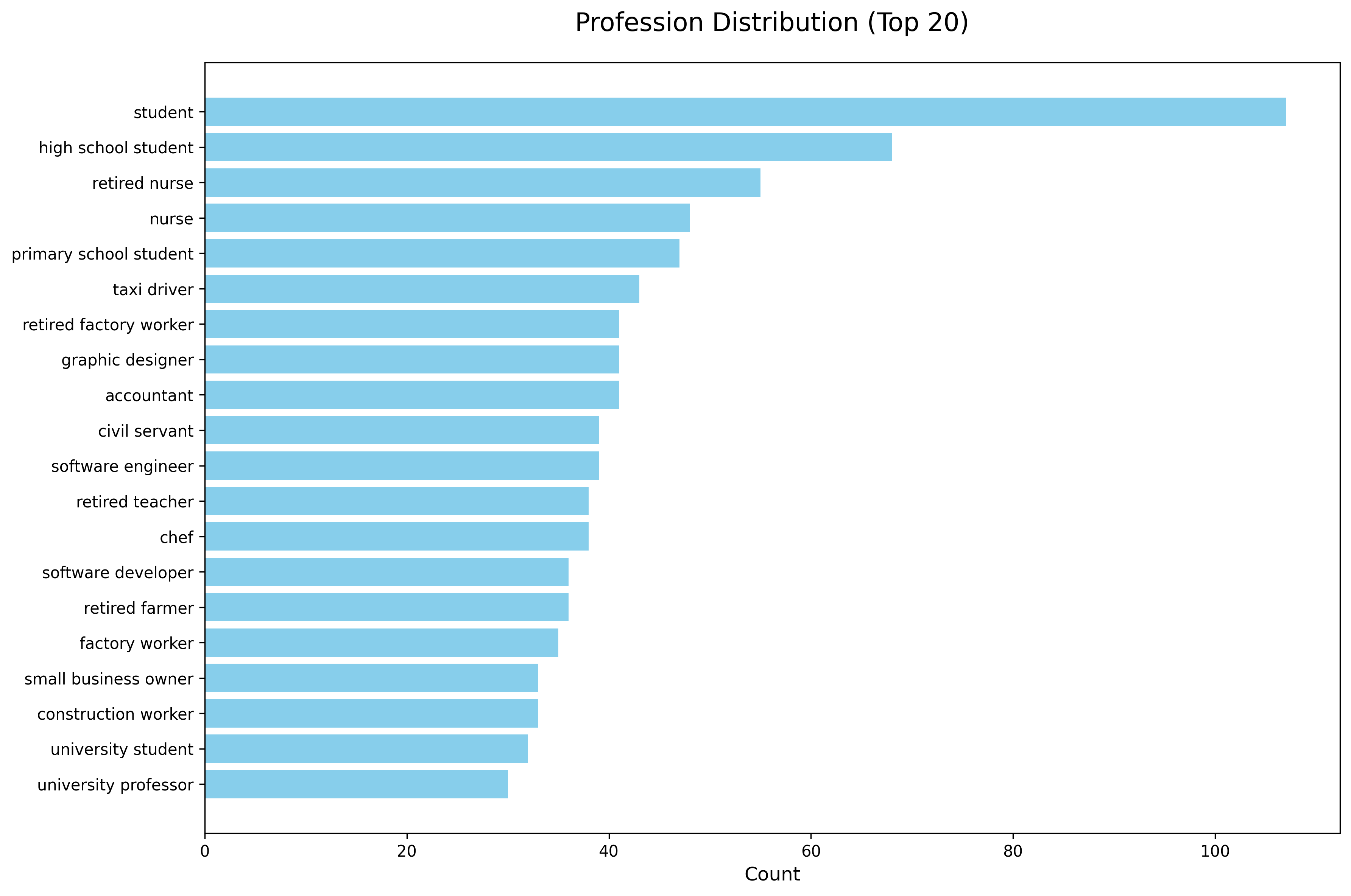}
        \caption{Top-20 most frequent professions among the personas.}
        \label{fig:persona_top20_profession}
    \end{subfigure}

    \vspace{0.6em}

    \begin{subfigure}[t]{0.46\linewidth}
        \centering
        \includegraphics[width=\linewidth]{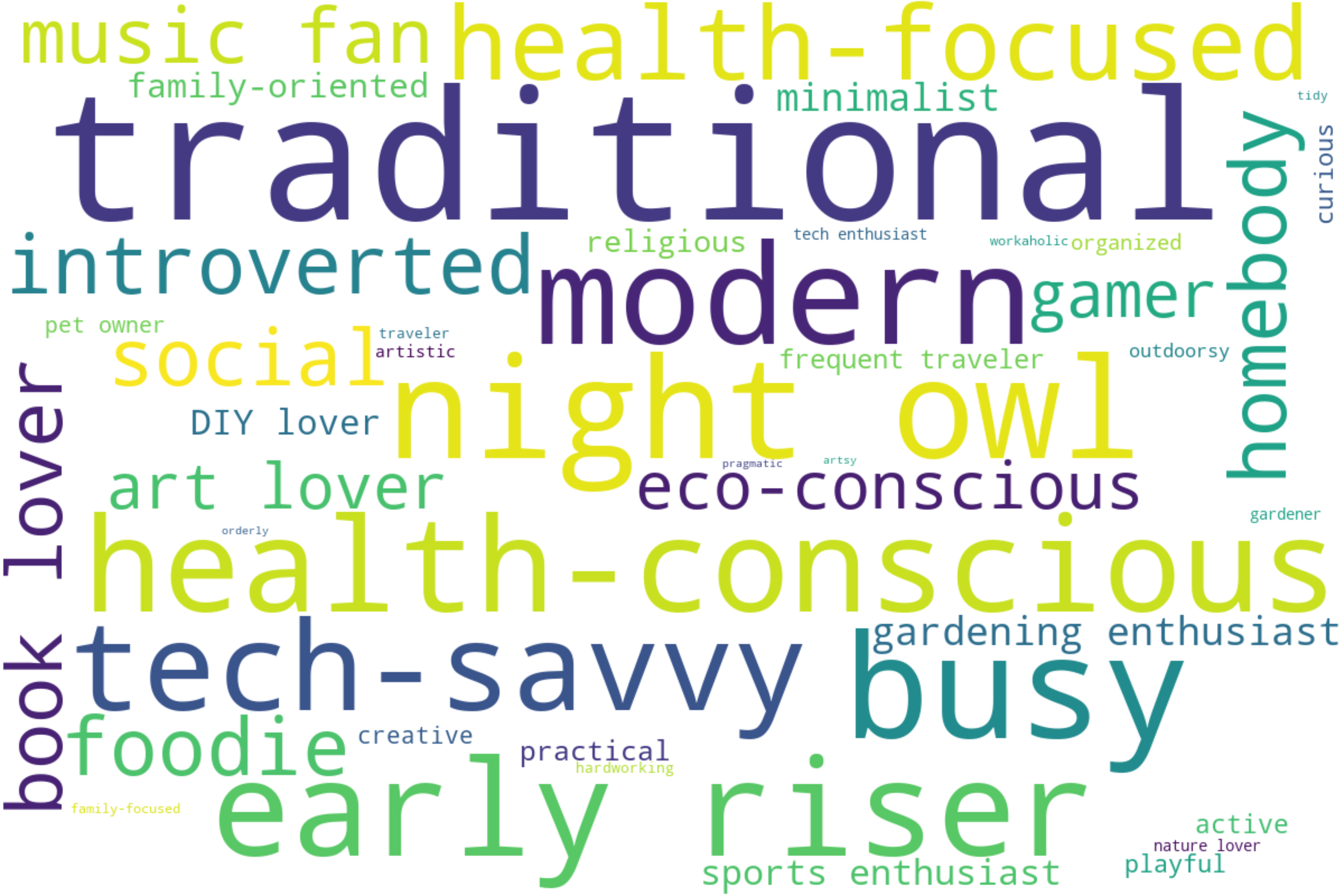}
        \caption{Word cloud of lifestyle tags describing common routines and living patterns.}
        \label{fig:persona_lifestyle}
    \end{subfigure}
    \hfill
    \begin{subfigure}[t]{0.46\linewidth}
        \centering
        \includegraphics[width=\linewidth]{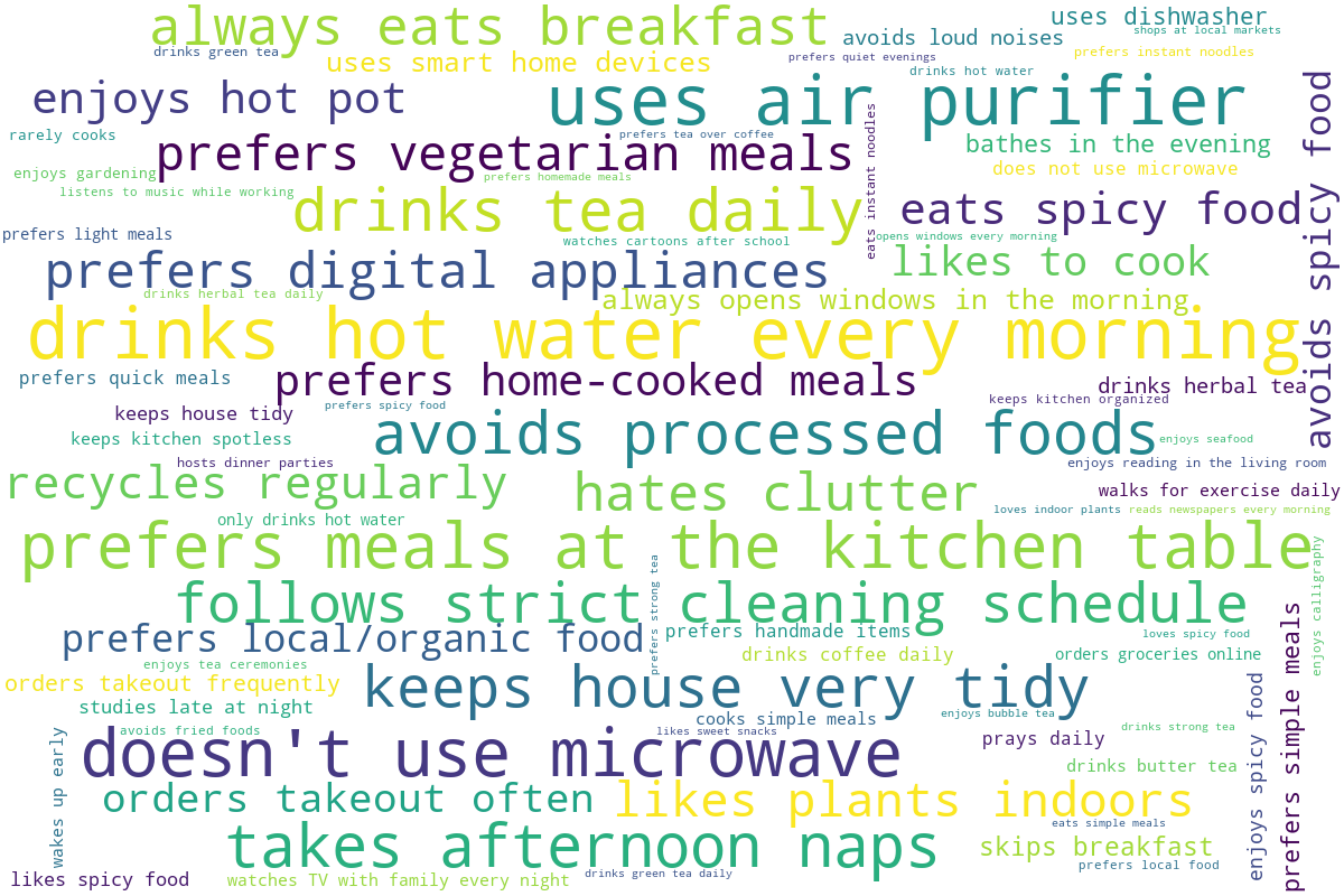}
        \caption{Word cloud of preference attributes reflecting diverse user needs and constraints.}
        \label{fig:persona_preferences}
    \end{subfigure}

    \caption{Statistical overview of the synthesized personas, including age distribution, profession distribution, lifestyle tags, and preference attributes.}
    \label{fig:persona_statistics_overview}
\end{figure}

\paragraph{Personas}
To model a broad user base, we generate 1,677 distinct personas spanning diverse demographics and preferences. As illustrated in Fig.~\ref{fig:persona_statistics_overview}, we present their age distribution, top-20 professions, and word-cloud visualizations of lifestyle tags and preference attributes. Crucially, these attributes guide our task synthesis, ensuring the generation of realistic, user-conditioned goals and constraints.

\paragraph{Tasks}
The final dataset contains 67.1k tasks that exhibit substantial structural complexity. Across the dataset, the average plan length is 10.8 steps. To quantify instruction abstractness, we define the abstraction ratio as the plan length divided by the instruction word count. Our dataset yields an average abstraction ratio of 0.76, demonstrating a high degree of implicit reasoning required. We provide representative task examples in Tables~\ref{tab:easy_task_examples}--\ref{tab:abstract_task_examples}, covering tasks with varying difficulty levels and degrees of abstractness. These tasks are generated using diverse prompt templates conditioned on scene graphs and sampled user personas. The complete set of task annotations is available in the released dataset.


\section{Additional Results and Analysis}\label{app:additional_results}

This appendix provides qualitative examples that illustrate how our method handles different instruction regimes, including \emph{easy} (explicit), \emph{complex} (long-horizon with multiple objectives), \emph{abstract} (high ambiguity), and \emph{human-authored} free-form requests. An overview of the specific tasks and their associated categories used in this analysis is summarized in Table~\ref{tab:appendixE_task_examples}. Subsequently, for each example, we report the model reply in three parts: task decomposition, subgoal grounding, and the final PDDL action sequence.

\begin{table*}[!h]
\centering
\caption{Task examples used in the qualitative analysis. We report the instruction and the associated qualitative category. Detailed decomposition, grounding, and plans are shown in subsequent listings.}
\label{tab:appendixE_task_examples}
\small
\setlength{\tabcolsep}{6pt}
\begin{tabular}{l p{0.78\linewidth}}
\toprule
\textbf{Category} & \textbf{Instruction} \\
\midrule
Easy &
I'm in the bathroom, can you bring me a towel? \\
Complex &
Collect scissors, tape, and colored pencils from the closet, office, and bedrooms. Set up all materials on the kitchen table for a family craft night, then power on the multiport hub in the living room so everyone can charge their devices during the activity. Afterwards, clean up by putting all used materials back in their original rooms. \\
Abstract &
Create a classy dinner atmosphere and tidy afterward \\
Human Task &
Give me some tea. \\
\bottomrule
\end{tabular}
\end{table*}

\subsection{Easy Task Example}\label{app:easy_example}
We first show an explicit request with a clear target object and destination. The output demonstrates a short decomposition with straightforward navigation, pickup, and handover. Details are provided on the project page: \textit{(\href{https://tgpo2026.github.io/TGPO/appendix_materials/task_examples/Easy_task_example.md}{View Example})}

\subsection{Complex Task Example}\label{app:complex_example}

Next we consider a long-horizon instruction that requires (i) collecting multiple items from different rooms, (ii) arranging them on a target surface with relative placement constraints, (iii) operating an appliance, and (iv) restoring items back to their original locations. This example highlights how the decomposition maintains a coherent \emph{setup $\rightarrow$ operate $\rightarrow$ cleanup} structure.
Details are provided on project page: 
\textit{(\href{https://tgpo2026.github.io/TGPO/appendix_materials/task_examples}{View Example})}

\subsection{Abstract Task Example}\label{app:abstract_example}

We show a high-level, underspecified instruction. The reply operationalizes ``classy dinner atmosphere'' into a concrete table-setting procedure by selecting and arranging representative dining items. Details are provided on project page: \textit{(\href{https://tgpo2026.github.io/TGPO/appendix_materials/task_examples}{View Example})}

\subsection{Human Task Example}\label{app:human_example}

We also include a human-authored free-form request. The model decomposes the instruction into a sequence of object manipulation and appliance operations, and grounds each subtask into PDDL subgoals. Details are provided on project page: \textit{(\href{https://tgpo2026.github.io/TGPO/appendix_materials/task_examples}{View Example})}

\section{Predefined PDDL Domain}\label{nips_app:pddl_domain}
We predefine a large-scale PDDL domain for household service robots, consisting of 28 predicates and 24 actions. Full domain details are provided on the project page: 
\textit{(\href{https://tgpo2026.github.io/TGPO/appendix_materials/pddl_domain}{View PDDL Domain})}

\clearpage

\end{document}